\documentclass{article}

\PassOptionsToPackage{numbers, compress}{natbib}
\usepackage[preprint]{neurips_2021}
\usepackage[utf8]{inputenc} % allow utf-8 input
\usepackage[T1]{fontenc}    % use 8-bit T1 fonts
\usepackage{hyperref}       % hyperlinks
\usepackage{url}            % simple URL typesetting
\usepackage{booktabs}       % professional-quality tables
\usepackage{amsfonts}       % blackboard math symbols
\usepackage{nicefrac}       % compact symbols for 1/2, etc.
\usepackage{microtype}      % microtypography
\usepackage{xcolor}         % colors
\usepackage{times}
\usepackage{latexsym}
\usepackage{microtype}
\usepackage{graphicx}
\usepackage{booktabs}  
\usepackage{multirow}
\usepackage{subfigure}
\usepackage{verbatim}
\usepackage{CJKutf8} 
\usepackage{amssymb}
\usepackage{amsmath}

\usepackage{algorithm}
\usepackage{algorithmic}
\usepackage{makecell}
 \usepackage{wrapfig}
\title{Rethinking InfoNCE: How Many Negative Samples Do You Need?}

\author{
Chuhan Wu$^\dagger$~~~~Fangzhao Wu$^\ddagger$~~~~Yongfeng Huang$^\dagger$\\
  $^\dagger$Department of Electronic Engineering \& BNRist, Tsinghua University, Beijing 100084, China \\
  $^\ddagger$Microsoft Research Asia, Beijing 100080, China \\ 
  \texttt{\{wuchuhan15,wufangzhao\}@gmail.com, yfhuang@tsinghua.edu.cn} \\
}

\begin{document}

\maketitle

\begin{abstract}

InfoNCE loss is a widely used loss function for contrastive model training. 
It aims to estimate the mutual information  between a pair of variables by discriminating between each positive pair and its associated $K$ negative pairs.
It is proved that when the sample labels are clean, the lower bound of  mutual information estimation is tighter when more negative samples are incorporated, which usually yields better model performance.
However, in many real-world tasks the labels often contain noise, and incorporating too many noisy negative samples for model training may be suboptimal.
In this paper, we study how many negative samples are optimal for InfoNCE in different scenarios via a semi-quantitative theoretical framework.
More specifically, we first propose a probabilistic model to analyze the influence of the negative sampling ratio $K$ on training sample informativeness.
Then, we design a training effectiveness function to measure the overall influence of training samples on model learning based on their informativeness.
We estimate the optimal negative sampling ratio using the $K$ value that maximizes the training effectiveness function.
Based on our framework, we further propose an adaptive negative sampling method that can dynamically adjust the negative sampling ratio to improve InfoNCE based model training.
Extensive experiments on different real-world datasets show our framework can accurately predict the optimal negative sampling ratio in different tasks, and our proposed adaptive negative sampling method can achieve better performance than the commonly used fixed negative sampling ratio strategy.

\end{abstract}

\section{Introduction}
InfoNCE~\cite{oord2018representation} loss is a popular choice of loss function for contrastive learning, which aims to maximize a lower bound of the mutual information between a pair of variables~\cite{he2020momentum}.
It is widely used in various fields like language modeling~\cite{chi2020infoxlm,sun2020contrastive}, search~\cite{huang2013learning,shao2020context} and recommendation~\cite{wu2019npa,sun2019bert4rec}, to learn discriminative deep representations.
In the InfoNCE framework, each positive sample is associated with $K$ randomly selected negative samples, and the task is typically formulated as a $K$+1-way classification problem, i.e., classifying which one is the positive sample~\cite{oord2018representation}.
In this way, the model is required to discriminate  the positive sample from the negative ones, which can help estimate the variable mutual information and learn discriminative representations~\cite{HjelmFLGBTB19}.
It is proved that if the sample labels are clean, a larger negative sampling ratio $K$ can lead to a tighter lower bound of variable mutual information~\cite{oord2018representation}, which usually yields better performance.
This is intuitive because more information of negative samples is exploited in model training.

Unfortunately, in many real-world tasks sample labels are not perfect and may contain noise~\cite{natarajan2013learning}.
For example, the non-clicked items are usually used as the negative samples in recommender system scenario.
However, some non-clicked items may match users' interest and they are not clicked due to many other reasons, e.g., shown in a low position.
They can be false negative sample for recommendation model training.
It is not suitable to incorporate too many noisy negative samples because they may lead to misleading gradients that are harmful to model learning~\cite{menon2019can}.
Thus, it is important to find the appropriate negative sampling ratio $K$ under different sample qualities to train accurate models.
However, existing studies on InfoNCE mainly focus on the selection of hard negative samples~\cite{behrmann2020unsupervised,robinson2020contrastive,kalantidis2020hard}, while the study on the choice of negative sampling ratio is very limited.

In this paper, we study the problem of how many negative samples are optimal for InfoNCE in different tasks via a semi-quantitative theoretical framework.
More specifically, we first propose a probabilistic model to analyze the influence of negative sampling ratio $K$ of InfoNCE on the informativeness of training samples.
Then, we propose a training effectiveness function  to semi-quantitatively measure the overall influence  of training samples on model learning based on their informativeness.
We use the $K$ value that maximizes this function to estimate the optimal negative sampling ratio for InfoNCE.
Based on our framework, we further propose an adaptive negative sampling \textit{(ANS)} method   that can dynamically adjust the negative sampling ratio to improve model training based on the characteristics of different model training stages.
We conduct experiments on multiple real-world datasets.
The results show that our framework can effectively estimate the optimal negative ratio in different tasks, and  our proposed adaptive negative sampling method \textit{ANS} can consistently achieve better performance than the commonly used negative sampling technique with fixed negative sampling ratio.

The main contributions of this paper include: 
\begin{itemize}
    \item We propose a semi-quantitative theoretical framework to analyze the influence of negative sampling ratio on InfoNCE and further estimate its optimal value.
    \item We propose an adaptive negative sampling method that can dynamically change the negative sampling ratio for different stages of model training.
    \item We conduct extensive experiments to verify the validity of our theoretical framework and the effectiveness of the proposed adaptive negative sampling method for InfoNCE based model training.
\end{itemize}
\section{Related Work}\label{sec:RelatedWork}
\subsection{InfoNCE Loss Function}
InfoNCE~\cite{oord2018representation} is a widely used loss function for contrastive learning.
It aims to estimate a lower bound of the mutual information between two variables.
A relevance function $f(\cdot, \cdot)$ is used to measure the non-normalized mutual information score  between them.
%It can be implemented by various functions like dot product~\cite{huang2013learning}.
For each positive sample $(x^+,c)$, it is associated with $K$ random negative samples, which are denoted as $[(x^-_1, c), (x^-_2, c), ..., (x^-_K, c)]$.
Then, the InfoNCE loss function $\mathcal{L}_K$ is formulated as follows:
\begin{equation}
    \mathcal{L}_K=-\log(\frac{e^{f(x^+,c)}}{e^{f(x^+,c)}+\sum_{i=1}^K e^{f(x^-_i,c)}}).
\end{equation}
According to~\cite{oord2018representation}, if the labels are fully reliable, the lower bound of the mutual information between $x^+$ and $c$ estimated by InfoNCE is formulated as:
\begin{equation}
    I(x^+,c)\geq \log(K+1)-\mathcal{L}_K.
\end{equation}
We can see that this lower bound is tighter when the number of negative samples $K$ is higher, which can usually lead to better performance of the models trained based on InfoNCE loss~\cite{he2020momentum}.

\subsection{Applications of InfoNCE}

InfoNCE has wide applications in many fields like language modeling~\cite{kong2019mutual,sun2020contrastive,chi2020infoxlm}, search~\cite{huang2013learning,chang2020pre} and recommendation~\cite{wu2019npa,wu2019nrms,sun2019bert4rec}.
For example, \citet{huang2013learning} proposed a deep structured semantic model (DSSM) for document retrieval.
For each pair of query and clicked document (regarded as positive sample),  they randomly sampled 4 unclicked documents as negative samples, and used the InfoNCE loss to train the model by optimizing the log-likelihood of the posterior click probability of the positive sample.
\citet{wu2019npa} proposed a personalized news recommendation method named NPA.
In this method, they regarded each clicked news as a positive sample, and randomly sampled 4 non-clicked news displayed in the same impression as negative samples.
The model was trained via the InfoNCE framework in a similar way.
\citet{chi2020infoxlm} proposed a multilingual pre-trained language model named InfoXLM.
They incorporated the InfoNCE framework into several self-supervion tasks like  multilingual masked language modeling and  cross-lingual contrastive learning.
They used the momentum contrast~\cite{he2020momentum} method to construct a queue of negative samples with a size of 131,072 for model training.
In these methods, the negative sampling ratio is usually empirically selected, which requires heavy hyper-parameter search.
In addition, these methods leverage a fixed negative sampling ratio, which may not be optimal for different stages of model training. 
Our framework presented in this paper can help estimate the optimal negative sampling ratio in different scenarios.
In addition, our proposed adaptive negative sampling method \textit{ANS} can dynamically adjust the negative sampling ratio according to the characteristics of different model training stages.
Our approach has the potential to benefit many tasks that involve InfoNCE loss for model training.

\section{Our Framework}\label{sec:Model}

\begin{wrapfigure}{R}{0pt}
	\centering 
	\subfigure[Real.]{
	\includegraphics[width=0.3\textwidth]{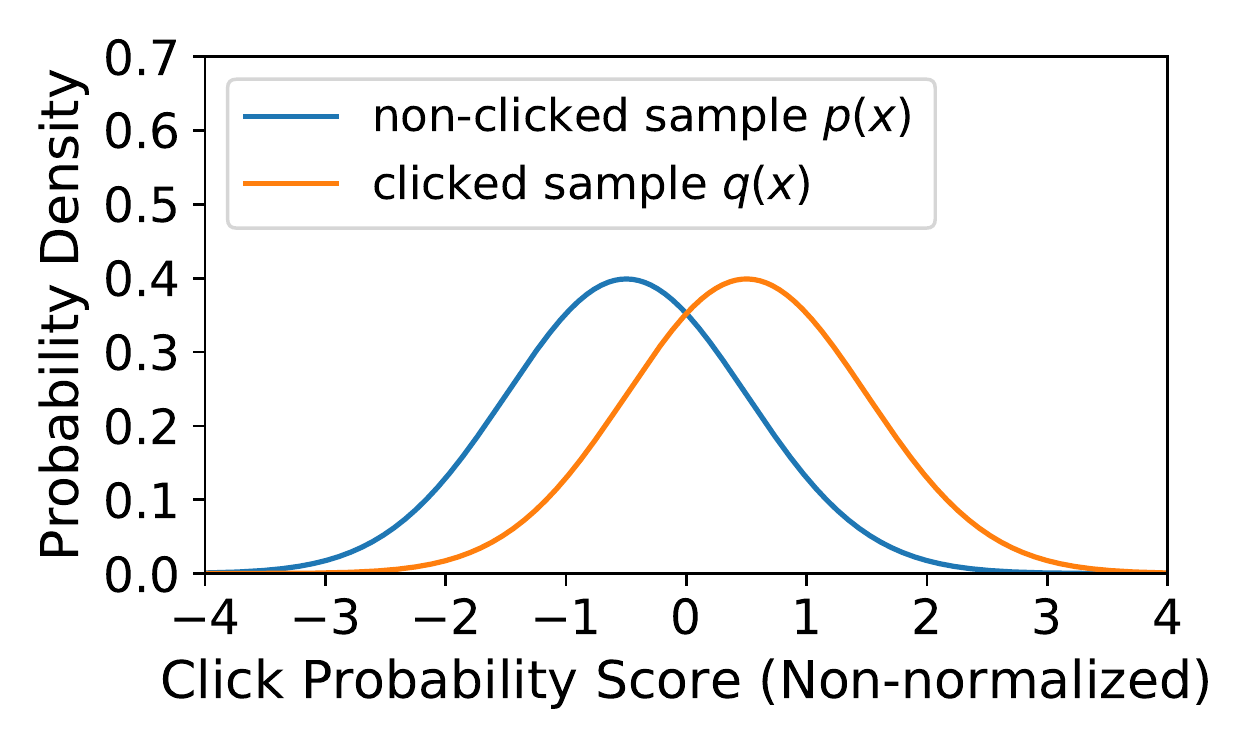} \label{fig.realdis}
	}
\quad
 	\subfigure[Predicted.]{
	\includegraphics[width=0.3\textwidth]{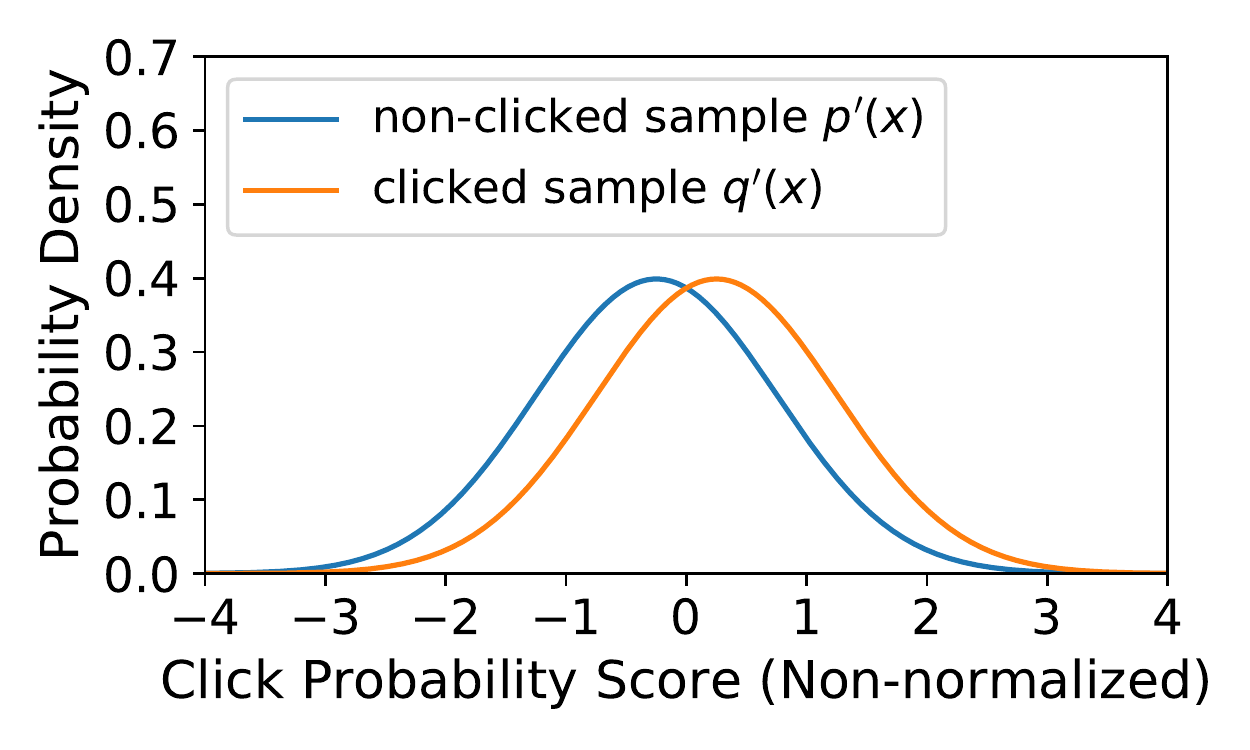} \label{fig.predictdis}
	}
\caption{Illustrated click probability score distributions of real positive and negative samples and the predicted ones.}\label{fig.dis}
\end{wrapfigure}

In this section, we introduce our semi-quantitative theoretical framework to analyze the influence of negative sampling ratio on the performance of the models trained based on InfoNCE loss.
We take news recommendation as an example scenario to better explain our framework.
In this scenario, it is a common practice to regard the clicked news as positive samples and non-clicked news as negative ones~\cite{wu2019npa}.
A recommendation model is trained with the binary click  labels to predict whether a user will click a candidate news.
We assume that there exists a click probability for each sample, which reflects the relevance between the candidate news and user interest.
We denote the click probability score distributions of clicked and non-clicked news as $q(x)$ and $p(x)$ respectively.
In addition, we denote the click probability score distributions of predicted clicked and non-clicked news as $q'(x)$ and $p'(x)$.
Fig.~\ref{fig.dis} shows an example.
We expect the click probability scores of clicked news to be higher than non-clicked ones. 
However, since users' click behaviors have rich randomness, there are overlaps between the click probability curves of positives and negatives, which means that the labels are noisy.
In addition, we assume the real labels are more discriminative (the curve overlap is smaller) than the predicted ones, which means the prediction accuracy is limited by the label quality.

Next, we introduce several key concepts in our framework.
Given a user $u$, a positive news $x^+$ and its associated $K$ negative news $[x^-_1, x^-_2, ..., x^-_K]$ are combined as a training sample.
We denote their click probability scores as $y^+$ and $[y^-_1, y^-_2, ..., y^-_K]$, respectively.
If they satisfy $y^+>\max(y^-_1, y^-_2, ..., y^-_K)$, we call the label of this training sample ``reliable''.
In a similar way, we denote the click probability scores of the predicted positive and negative news as $\hat{y}^+$ and $[\hat{y}^-_1, \hat{y}^-_2, ..., \hat{y}^-_K]$, respectively.
We regard the prediction for them as ``reliable'' if $\hat{y}^+>\max(\hat{y}^-_1, \hat{y}^-_2, ..., \hat{y}^-_K)$.
For simplicity, we assume that all the $K+1$ candidate news are independent.
We define events $A$ and $B$ as the ``reliability'' of label and model prediction respectively. 
Then, the probabilities $P(A)$ and $P(B)$ are formulated as follows:
\begin{equation}\small
    P(A)=\int_{-\infty}^{\infty}q(x)[\int_{-\infty}^{x}p(y)dy]^Kdx,
\end{equation}

\begin{equation}\small
    P(B)=\int_{-\infty}^{\infty}q'(x)[\int_{-\infty}^{x}p'(y)dy]^Kdx.
\end{equation}

Then, we introduce how to measure the informativeness of training samples and their influence on model learning.
If the model prediction on a sample is ``unreliable'' while the label is ``reliable'', then this sample is very informative for calibrating the model.
We regard this kind of samples as ``good samples'', and denote their set as $\mathcal{G}$.
On the contrary, if the model prediction on a sample is ``reliable'' but the label is ``unreliable'', this sample is harmful for model training because it will produce misleading gradients.
We regard this kind of samples as ``bad samples'', and  their set is denoted as $\mathcal{B}$.
For the rest samples, the model predictions and labels have the same ``reliability'', and these samples are usually less informative for model training.\footnote{When predictions and labels are both ``reliable'', the model can simply optimize the losses on these samples, which may lead to overfitting. When predictions and labels are both not ``reliable'', the gradients are also not very informative.}
We call this kind of samples as ``easy samples'', and their set is denoted as $\mathcal{E}$.
For simplicity, here we make an assumption that the events $A$ and $B$ are independent.
We denote the total number of training samples as $N$.
Then, the number of each kind of samples introduced above is formulated as follows:
\begin{equation}
\begin{aligned}
|\mathcal{G}|& =P(A)[1-P(B)]N,\\
|\mathcal{B}|& =P(B)[1-P(A)]N,\\
|\mathcal{E}|& =N-|\mathcal{G}|-|\mathcal{B}|.\\
\end{aligned}
\end{equation}
To semi-quantitatively measure the influence of training samples on model learning, we propose a training effectiveness metric $v$ based on the different informativeness of training samples, which is formulated as follows:
\begin{equation}
v= \frac{1}{N}[\lambda(|\mathcal{G}|-|\mathcal{B}|)+(1-\lambda)|\mathcal{E}|], \label{eqv}
\end{equation}
where $\lambda$ is a hyperparameter that controls the relative importance of good and bad samples.
According to our massive experiments, we find that setting $\lambda=0.9$ is appropriate for estimating the negative sampling ratio.
In Eq. (\ref{eqv}), given the distributions $p(x)$, $q(x)$, $p'(x)$ and $q'(x)$, $v$ is only dependent on the negative sampling ratio $K$.
Without loss of generality, we assume that these distributions are all Gaussian distributions with the same standard deviation 1, and the mean values of $p(x)$ and $p'(x)$ (denoted as $\mu_{p(x)}$ and $\mu_{p'(x)}$) satisfy $\mu_{p(x)}=\mu_{p'(x)}=0$.
We still need to estimate the mean values of $q(x)$ and $q'(x)$ (denoted as $\mu_{q(x)}$ and $\mu_{q'(x)}$).
In practice, we can estimate $\mu_{q(x)}$ using the training AUC score under $K=1$ before overfitting\footnote{We assume the model is strong enough to fit the training data.} by solving the following equation\footnote{We cannot obtain a closed-form solution of $\mu_{q(x)}$ due to the characteristic of Gaussian distribution. Thus, we need to solve $\mu_{q(x)}$ with numerical methods (e.g., bisection method).}:
\begin{equation}\small
    AUC= \int_{-\infty}^{\infty}\frac{1}{2\pi}e^{\frac{-(x-\mu_{q(x)})^2}{2}}[\int_{-\infty}^{x}e^{\frac{-y^2}{2}}dy]dx.
\end{equation}
In a similar way, the value of $\mu_{q'(x)}$ can be estimated based on the AUC score on the validation set under $K=1$.
When $\mu_{q(x)}$ and  $\mu_{q'(x)}$ are known, we can compute the current training effectiveness. 

Finally, we introduce how to estimate the optimal negative sampling ratio based on our proposed framework.
Since the model may produce different $p'(x)$ and $q'(x)$ during the training process, the training effectiveness measurement is time-variant.
Thus, we denote the training effectiveness value at the $i$-th iteration step as $v_i$, and the overall training effectiveness $v$ is computed as $
    v=\frac{1}{T}\sum_{i=1}^T v_i,$
where $T$ is the number of iterations needed for model convergence.
Since $v$ is the function of $K$ and other variables can be approximated, we can estimate the optimal value of $K$ that maximizes $v$.

\section{Experiments}\label{sec:Experiments}
In this section, we conduct experiments to verify our proposed framework.
We first introduce the datasets and experimental settings, and then introduce the results and findings.

\begin{wraptable}{R}{8.5cm}
\centering
\caption{Detailed dataset statistics.}\label{table.dataset}
 \vspace{0.1in}
	\resizebox{0.6\textwidth}{!}{
\begin{tabular}{lrlr}
\Xhline{1.5pt} 
\multicolumn{4}{c}{MIND}                                      \\ \hline
\# News         & 161,013    & \# User           & 1,000,000  \\
\# Impression   & 15,777,377 & \# Click behavior & 24,155,470 \\
Avg. title len. & 11.52      & Avg. body len.    & 585.05     \\ \hline
\multicolumn{4}{c}{ML-1M}                                     \\ \hline
\# Item         & 3,706      & \# User           & 6,040      \\
\# Interaction  & 1,000,209  & Avg. history len. & 165.6      \\  
\Xhline{1.5pt}
\end{tabular}
} 
%\vspace{-0.1in}
\end{wraptable}

\subsection{Datasets and Experimental Settings}

In our experiments, we verify our framework in three tasks, i.e., news recommendation, news title-body matching and item recommendation.
The first two tasks are performed on the MIND~\cite{wu2020mind} dataset\footnote{https://msnews.github.io/}, which contains the news impression logs of 1 million users in 6 weeks.
For the news recommendation task, each clicked news is regarded as a positive sample and the news displayed in the same impression but not clicked by the user are regarded as negative samples.
The logs in the last week are used for test, and the rest are used for training and validation.
For the news title-body matching task, we regard the original news title-body pairs as positive samples, and negative samples are created by randomly pairing titles and bodies.\footnote{Some randomly combined title-body pairs happen to be from the same news, i.e., they are actually positive samples. We remove these pairs from the negative sample set.}
We use the news in the training set of MIND for model training, and those in the validation and test sets (except the news included in the training set) for validation and test respectively.
The item recommendation task is performed on the MovieLens dataset~\cite{harper2015movielens}, and we use the ML-1M~\footnote{https://grouplens.org/datasets/movielens/1m/} version for experiments.
We use the same experimental settings as~\cite{sun2019bert4rec}.
The statistics of datasets are summarized in Table~\ref{table.dataset}.

In these experiments, we use NRMS~\cite{wu2019nrms} as the base model for news recommendation, a Siamese Transformer model~\cite{reimers2019sentence} for title-body matching, and use BERT4Rec~\cite{sun2019bert4rec} for item recommendation.\footnote{The detailed hyperparameter settings are in supplements.}
We use AUC and nDCG@10 to measure the performance of news and item recommendation, and use AUC and HR@5 as the metrics for news title-body matching.
All these models are trained with the InfoNCE loss under different negative sampling ratios.
Each experiment is repeated 5 times and the average performance is reported.

\subsection{Experimental Results}

\begin{figure*}[!t]
	\centering 
	\subfigure[Training and validation AUC.]{
	\includegraphics[height=1.2in]{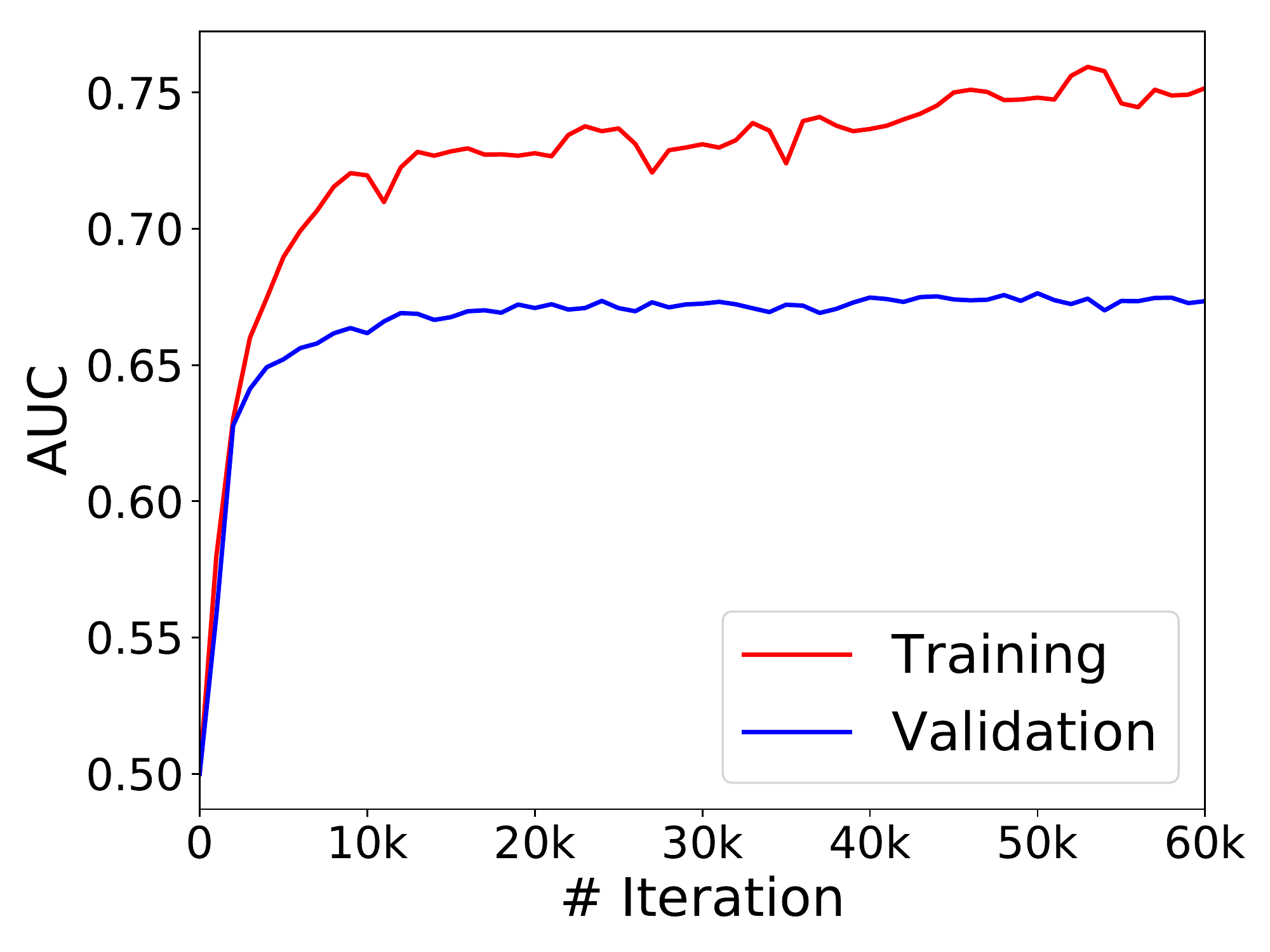} \label{fig.result11}}
 	\subfigure[Estimated overall training effectiveness $v$ under different $K$ and  $\lambda$.]{
	\includegraphics[height=1.2in]{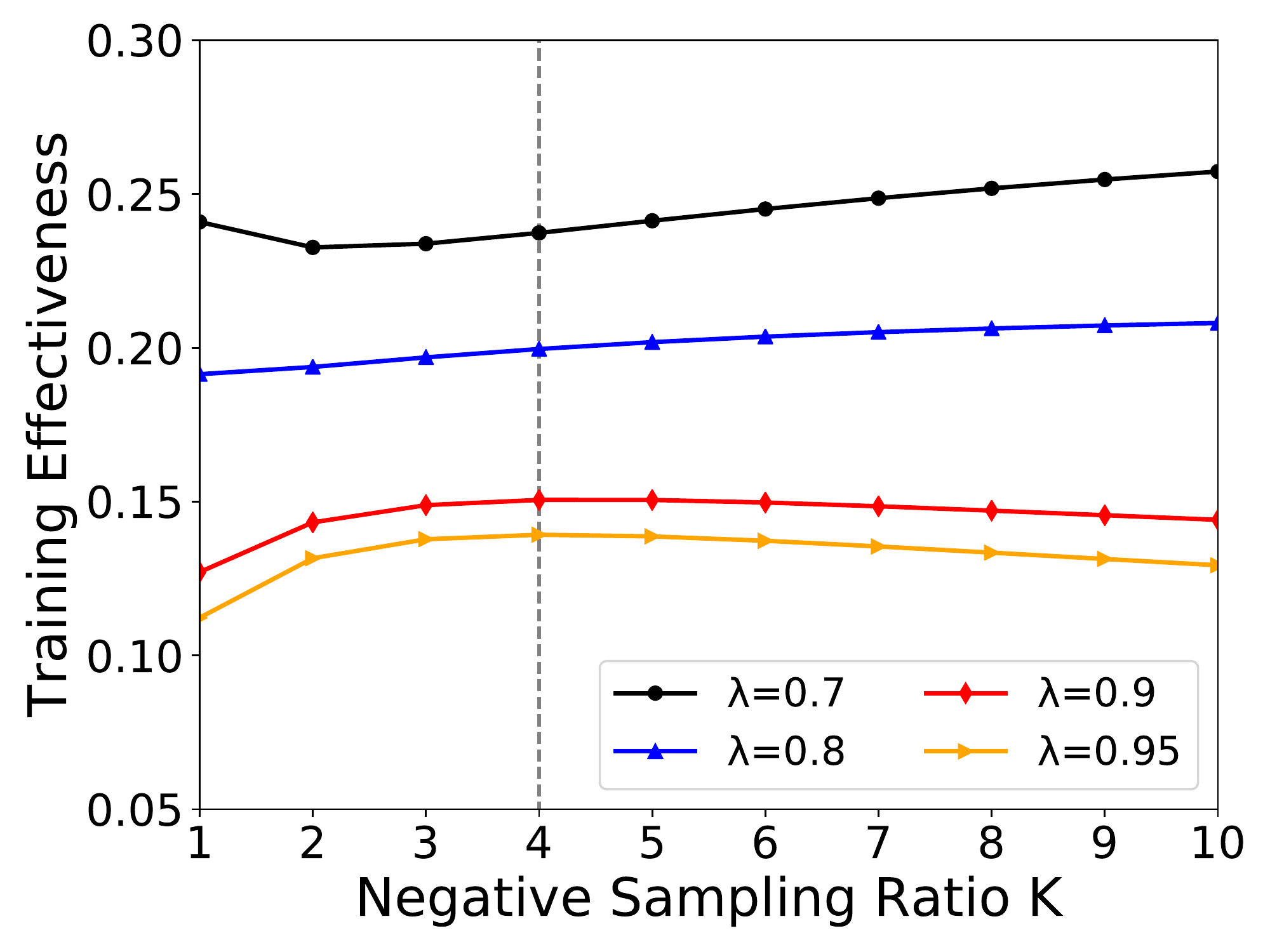}\label{fig.result12}}
	 	\subfigure[Performance under different $K$.]{
	\includegraphics[height=1.2in]{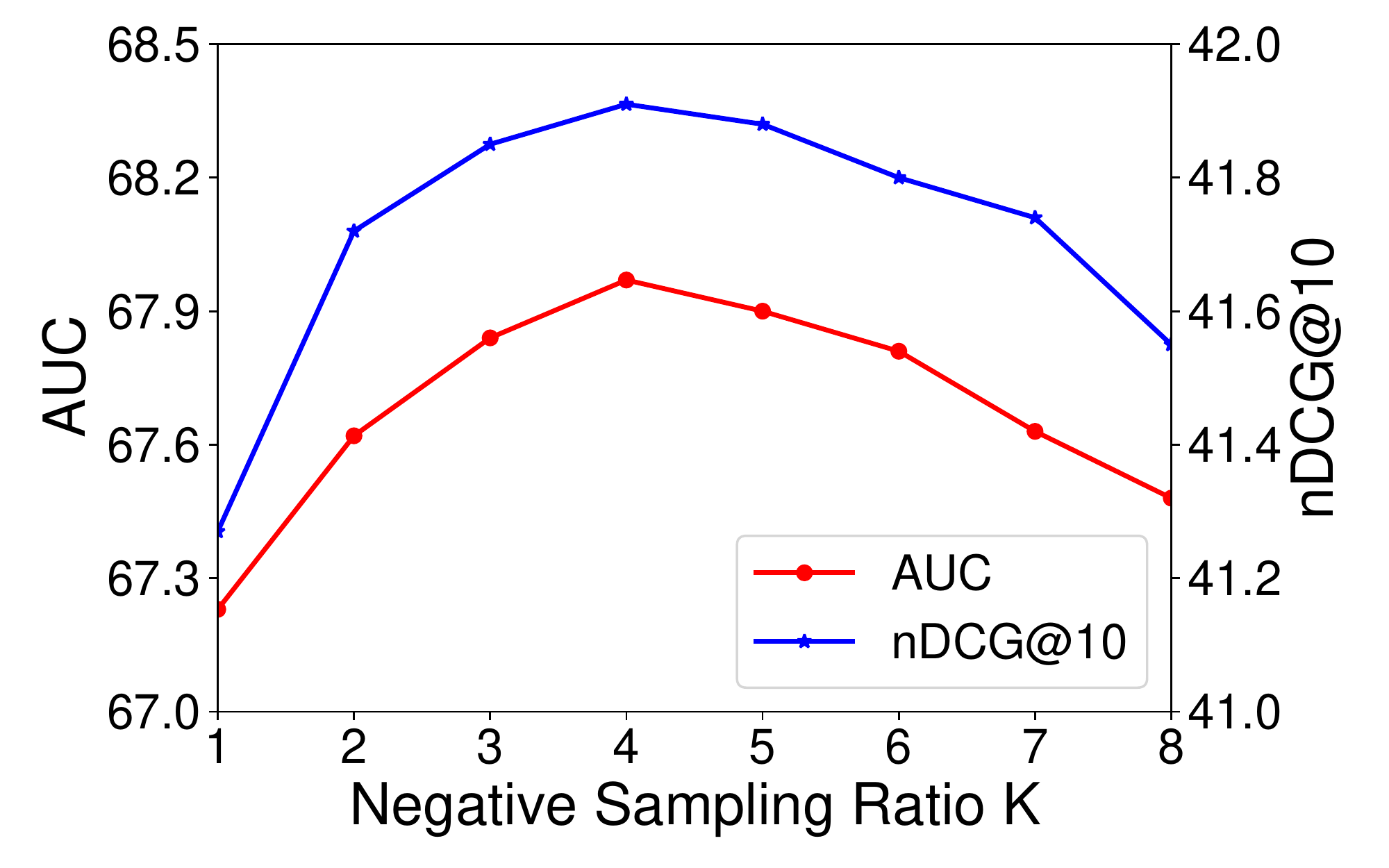}\label{fig.result13}} 
\caption{Results and predictions on the news recommendation task. Gray dashed line represents the optimal $K$ under $\lambda=0.9$.} \label{fig.result1}%\vspace{-0.015in}

\end{figure*}

\begin{figure*}[!t]
	\centering 
	\subfigure[Training and validation AUC.]{
	\includegraphics[height=1.2in]{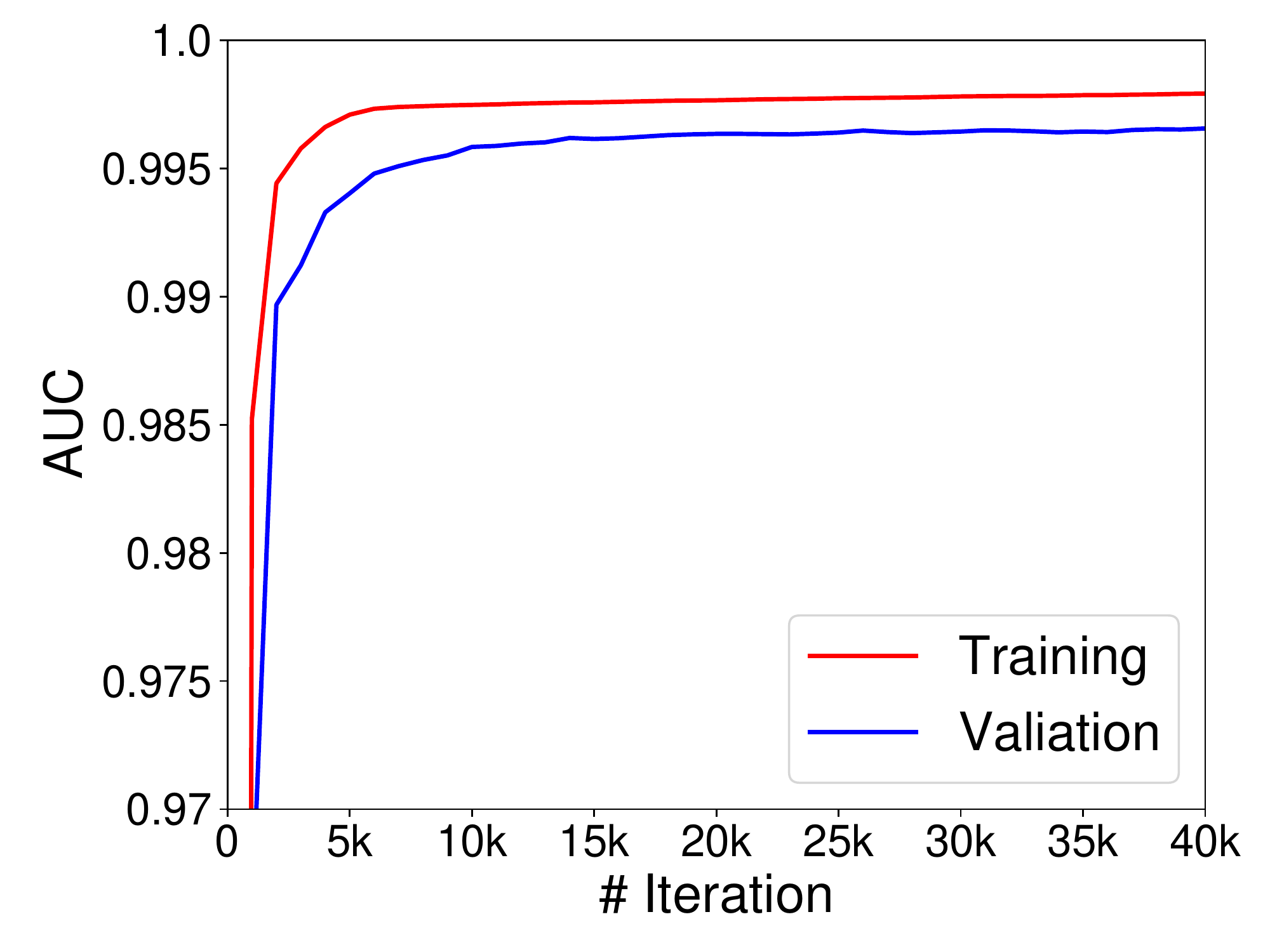} \label{fig.result21}}
 	\subfigure[Estimated overall training effectiveness $v$ under different $K$ and  $\lambda$. ]{
	\includegraphics[height=1.2in]{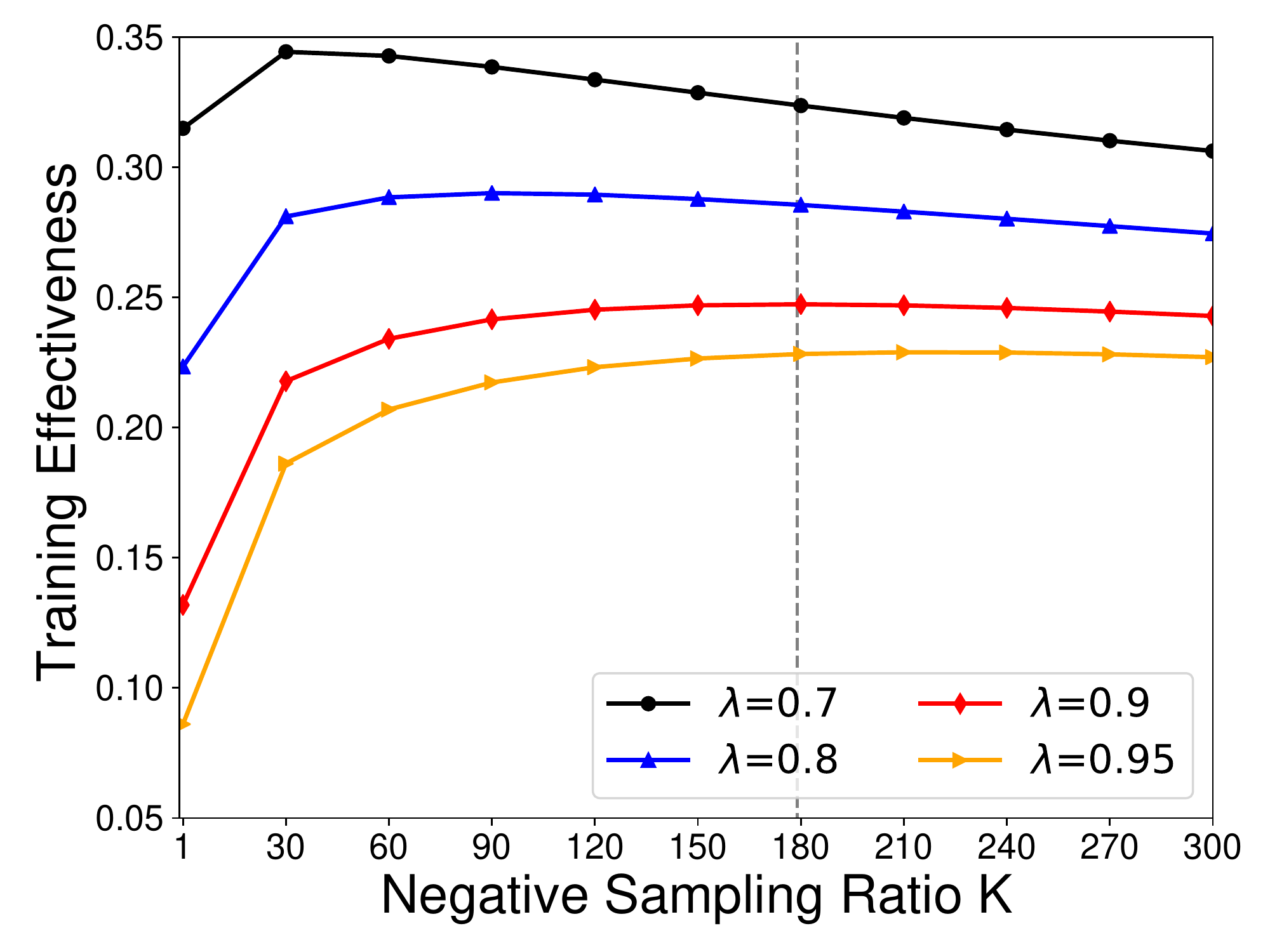} \label{fig.result22}}
	 	\subfigure[Performance under different $K$.]{
	\includegraphics[height=1.2in]{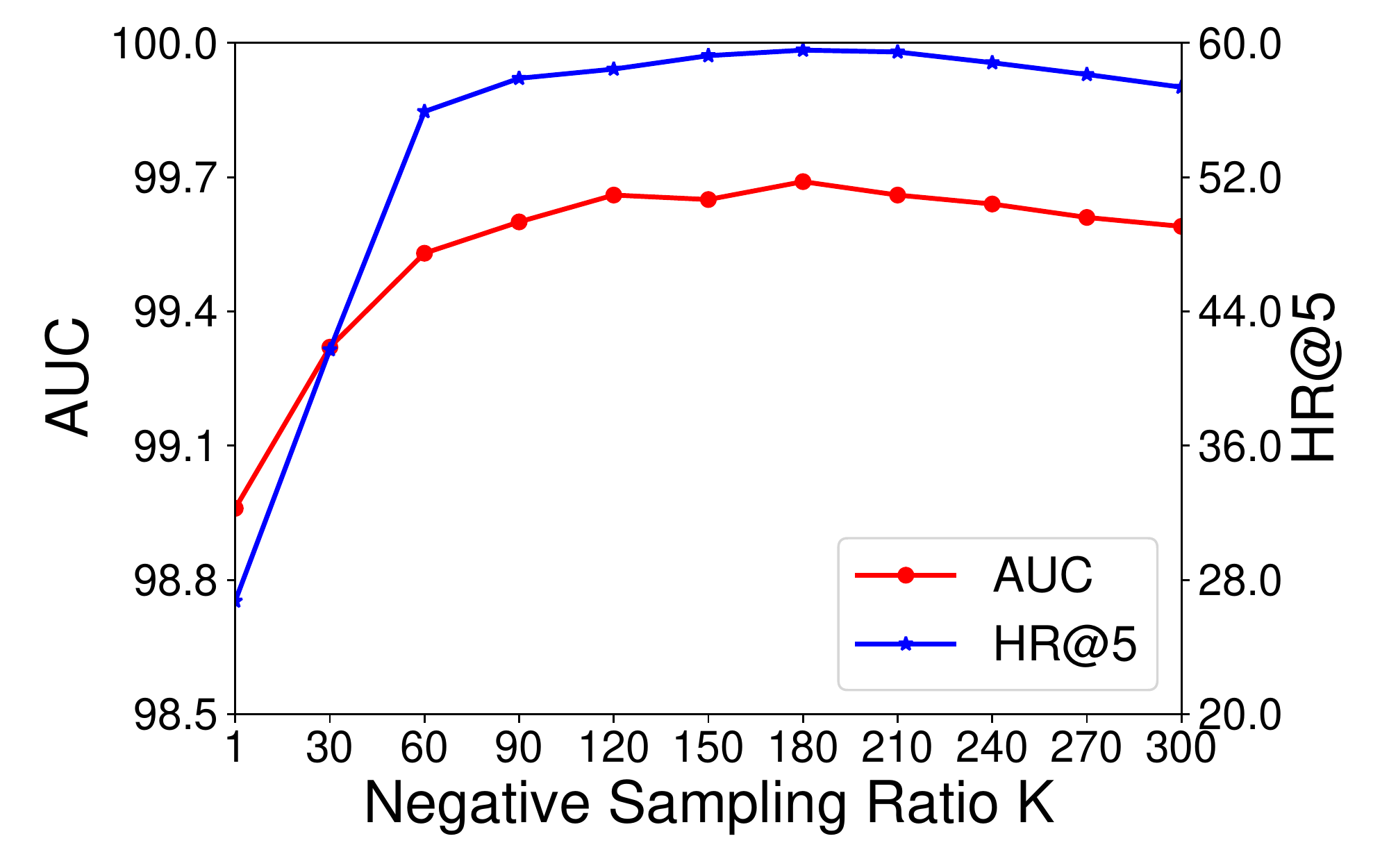} \label{fig.result23}}%\vspace{-0.05in}
\caption{Results and predictions on the title-body matching task. Gray dashed line represents the optimal $K$ under $\lambda=0.9$.} \label{fig.result2}%\vspace{-0.1in}
%\vspace{-0.1in}
\end{figure*}

\begin{figure*}[!t]
	\centering 
	\subfigure[Training and validation AUC.]{
	\includegraphics[height=1.2in]{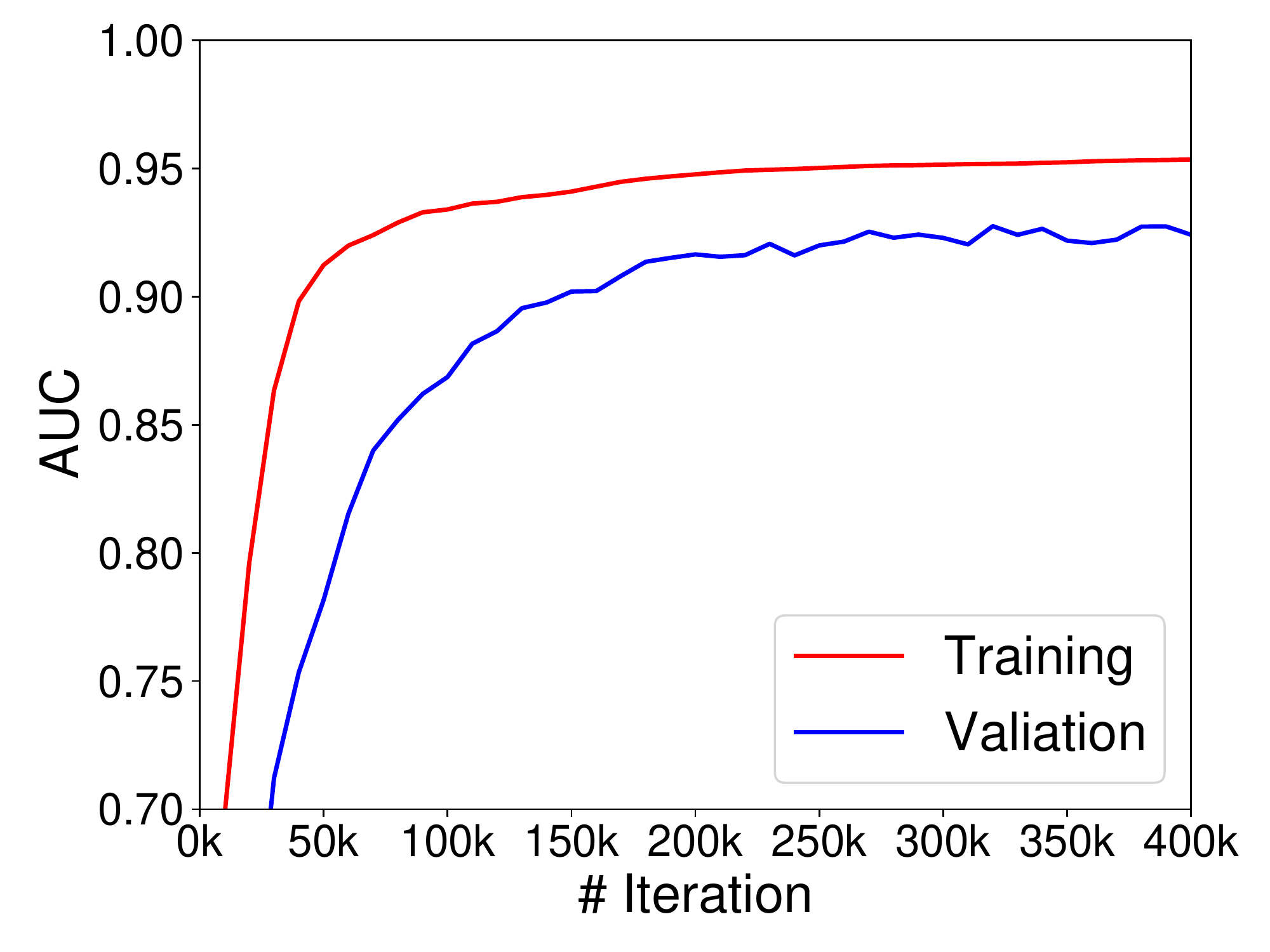} \label{fig.result31}}
 	\subfigure[Estimated overall training effectiveness $v$ under different $K$ and  $\lambda$. ]{
	\includegraphics[height=1.2in]{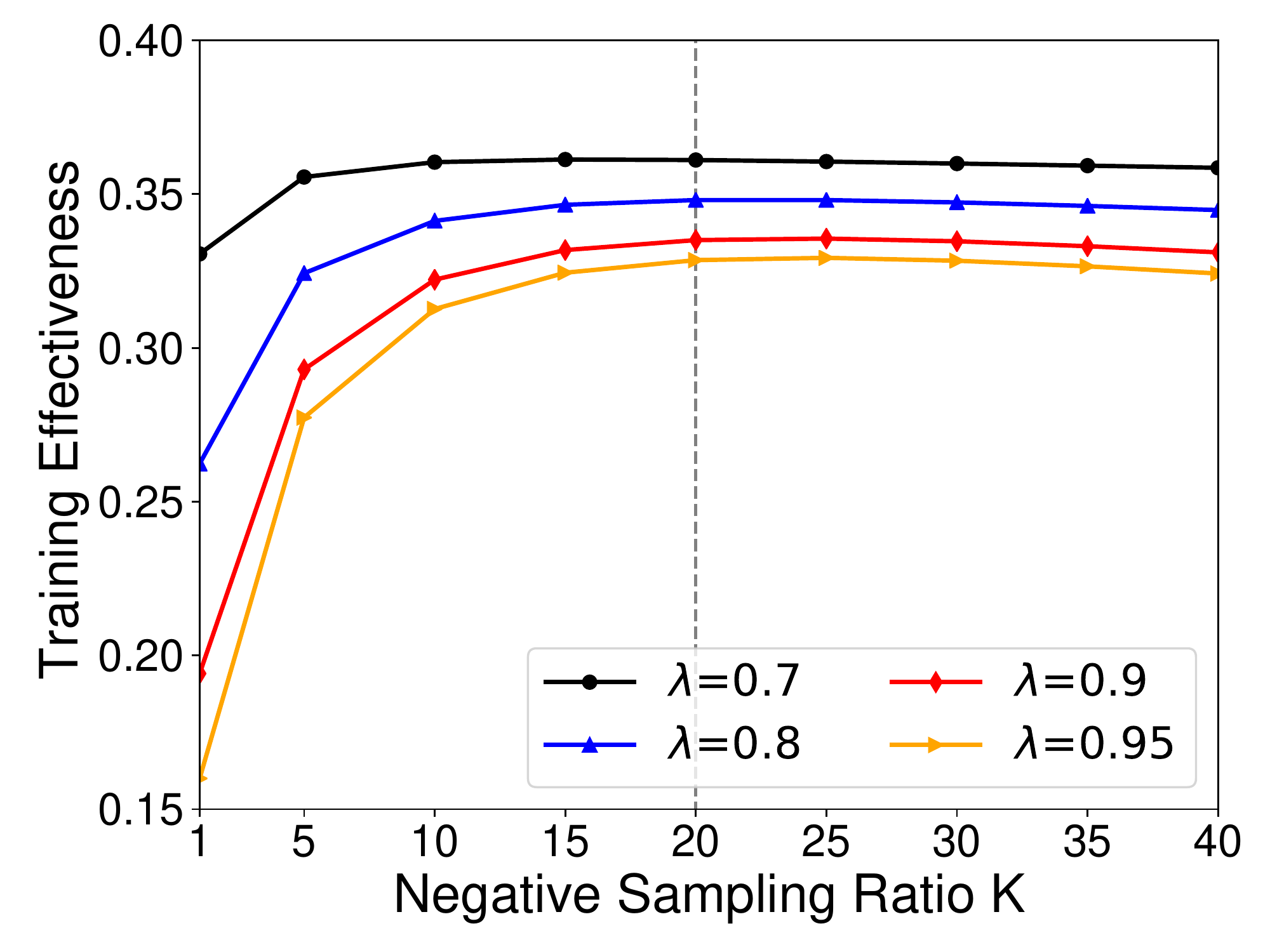} \label{fig.result32}}
	 	\subfigure[Model performance under different $K$.]{
	\includegraphics[height=1.2in]{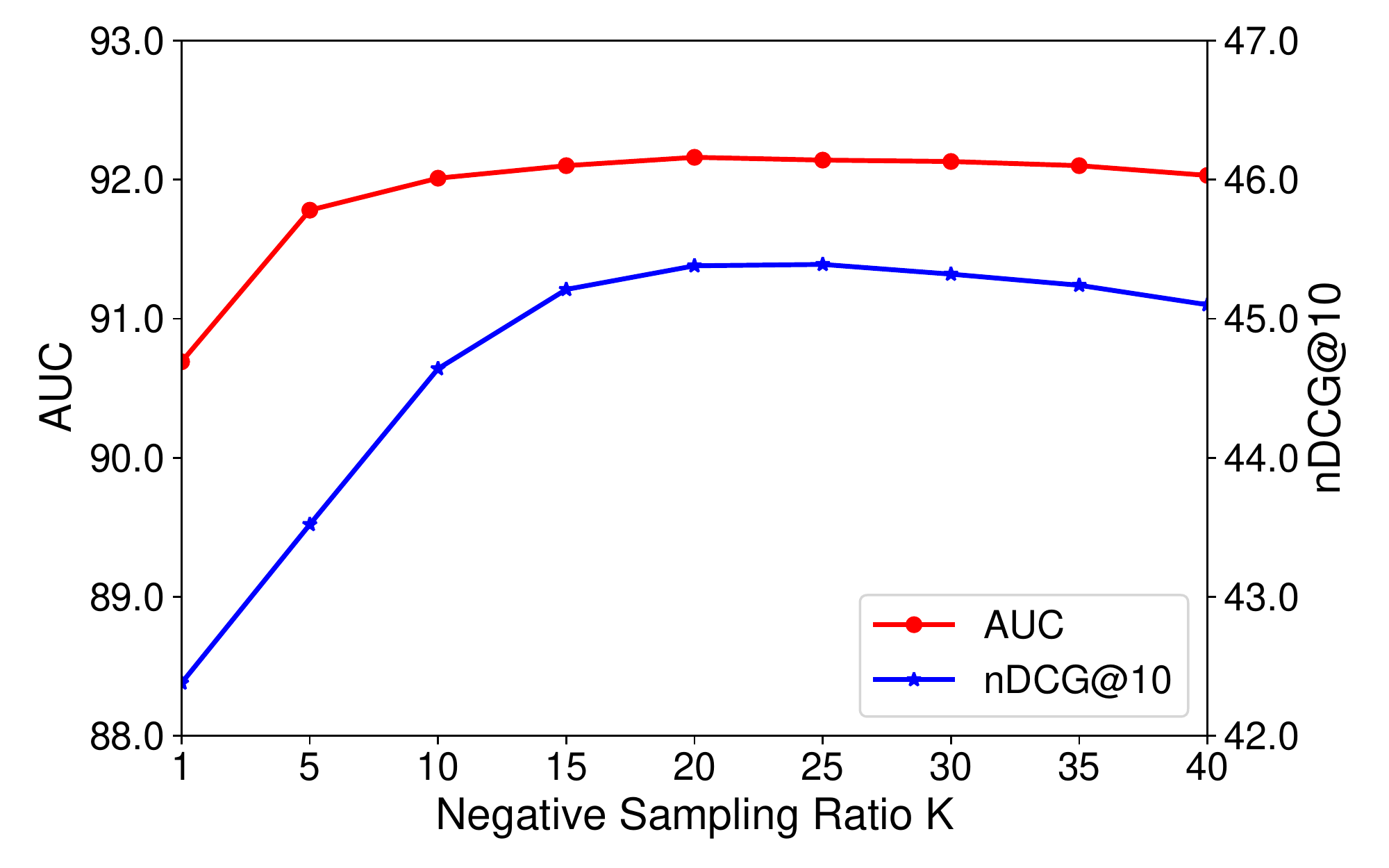} \label{fig.result33}}
\caption{Results and predictions on the movie recommendation task. Gray dashed line represents the optimal $K$ under $\lambda=0.9$.} \label{fig.result3}
\end{figure*}

In this section, we verify our framework according to the experiments on different tasks.
Fig.~\ref{fig.result1} shows the model training and validation AUC curves, the estimated overall training effectiveness under different negative sampling ratio $K$ and hyperparameter $\lambda$, and the real model performance under different negative sampling ratios $K$ in the news recommendation task.
From Fig.~\ref{fig.result11}, we can observe that the training AUC is around 0.75 when model converges, which means that $\mu_{q(x)}$ is about 1.
From Figs.~\ref{fig.result13}, we find that the real model performance is not optimal when $K$ is too small or too large, and $K=4$ yields the best performance, which is consistent with the red ($\lambda=0.9$) and orange ($\lambda=0.95$) curves in Fig.~\ref{fig.result12}.
It shows that the value of $\lambda$ should be relatively large, which is intuitive because the influence of good and bad samples on the model training is usually dominant.
Fig.~\ref{fig.result2} shows the results on the title-body matching task.
From Fig.~\ref{fig.result21}, we can estimate that the value $\mu_{q(x)}$ in this task is around 4.5 (training AUC is about 0.998).
We find it interesting that when $\lambda=0.9$, the estimated optimal negative sampling ratio ($K=180$) in Fig.~\ref{fig.result22} is consistent with the experimental results in Fig.~\ref{fig.result23}.
From Fig.~\ref{fig.result3}, we have similar findings on the item recommendation task.
The training AUC is about 0.95, which approximately corresponds to $\mu_{q(x)}=2.4$ in our theoretical framework. 
The estimated optimal negative sampling ratio $K$ is about 20, which is consistent with the experimental results in Fig.~\ref{fig.result33}.
Thus, setting the hyperparameter $\lambda$ to 0.9 would be appropriate for estimating the optimal negative sampling ratio.
In addition, from these figures we find that the model performance improves first when $K$ increases, and then starts to decline when $K$ becomes too large.
This is probably because useful information in negative samples cannot be exploited when $K$ is too small, while the label noise will harm the model training when $K$ is too large.
Thus, a medium value of $K$ may be more suitable for model training with noisy labels.

\subsection{Analysis}

\begin{figure}[t]
	\centering 
	
	\includegraphics[width=0.5\textwidth]{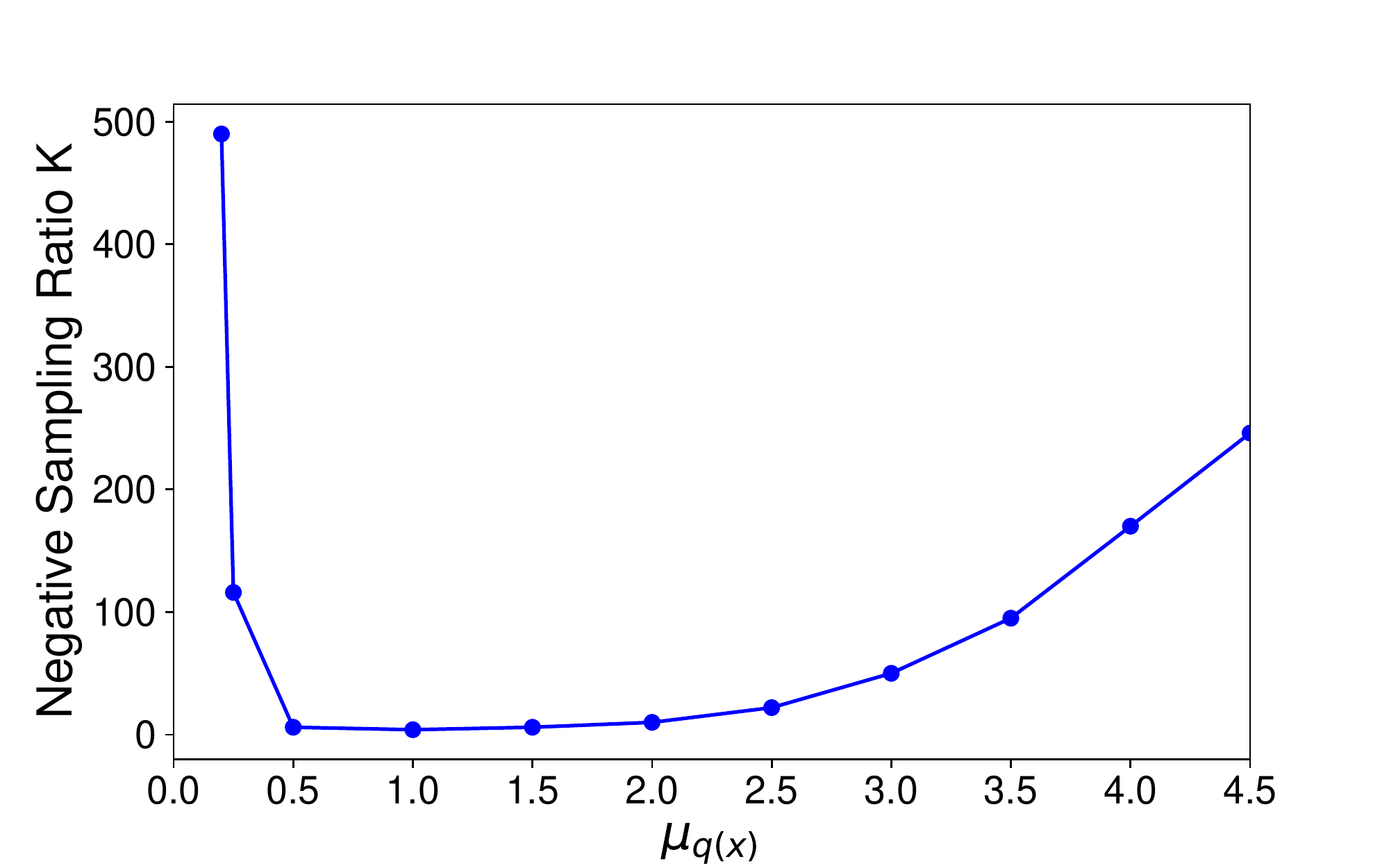}%\vspace{-0.05in}
\caption{The simulated optimal value of $K$ under different $\mu_{q(x)}$. }%\vspace{-0.1in} 
\label{fig.negsim}
\end{figure}

In this section, we use our framework to further analyze the negative sampling ratio in InfoNCE.
We set the value of $\lambda$ to 0.9 in the following analysis.
First, we study the influence of the label quality  (represented by $\mu_{q(x)}$) on the estimated optimal value of $K$.
We use the function $\mu_{q'(x)}^t=\mu_{q(x)}(1-e^{-t}), t\in[0, 3]$ to simulate the model training curve\footnote{We find the shape of this curve is similar to the real training curves.}, and the estimated optimal values $K$ under different $\mu_{q(x)}$ are shown in Fig.~\ref{fig.negsim}.
We find it interesting that the optimal value of $K$ boosts when $\mu_{q(x)}$ is close to 0 or relatively large.
This may be because when $\mu_{q(x)}$ is very small, we need many negative samples to counteract the noise.
And when $\mu_{q(x)}$ is large, we may also need increase the number of negative samples because the labels are relatively reliable.

\begin{figure*}[!t]
	\centering 
	
	\includegraphics[width=0.93\textwidth]{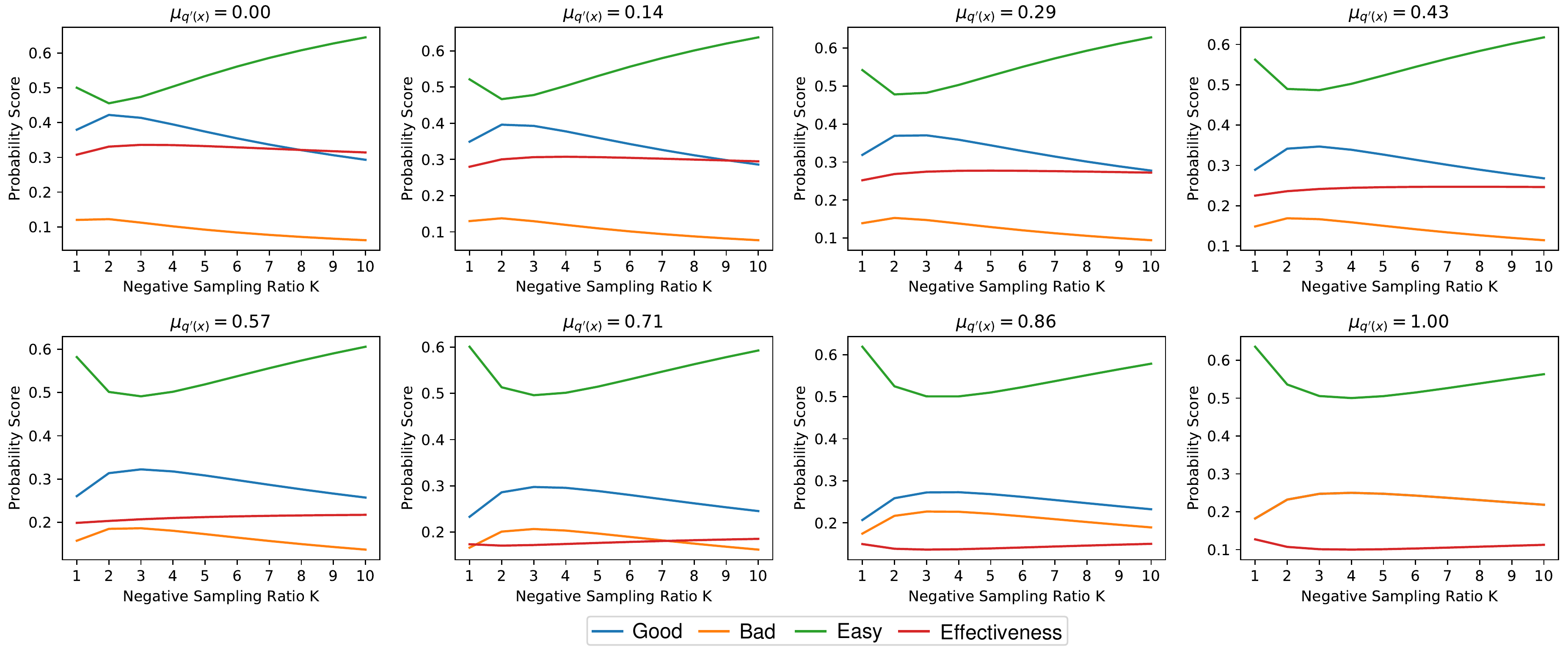}%\vspace{-0.05in}
\caption{Training effectiveness and proportion of different kinds of samples in different training stages on the news recommendation task. Good and bad curves overlap  in the last plot.} \label{fig.train1}
\end{figure*}

\begin{figure*}[!t]
	\centering 
	
	\includegraphics[width=0.93\textwidth]{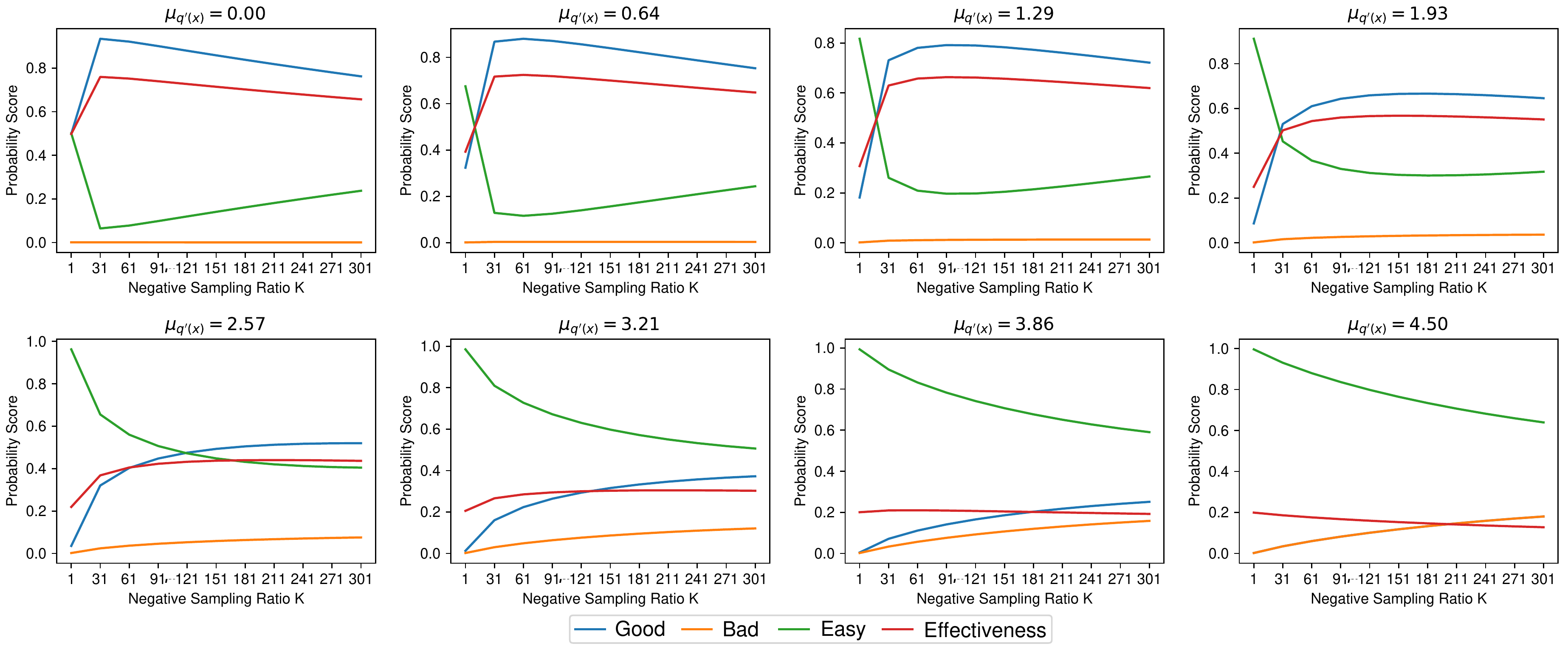}%\vspace{-0.05in}
\caption{Training effectiveness and proportion of different kinds of samples in different training stages on the title-body matching task. Good and bad curves overlap  in the last plot.} \label{fig.train2}
\end{figure*}

Next, we study how the training effectiveness and the proportions of different kinds of samples change at model training stages (represented by different $\mu_{q'(x)}$). 
The results on the news recommendation and title-body matching tasks are shown in Figs.~\ref{fig.train1} and~\ref{fig.train2}, respectively.\footnote{Due to space limit, we only present the results of news recommendation and title-body matching because the results in the item recommendation task show similar patterns.}
We have two interesting findings from them.
First, the optimal negative sampling ratio at the beginning of model training is smaller than the  globally optimal one estimated in the previous section.
This may be because when model is not discriminative, the ratio of good samples is much larger than bad samples, and using too many negative samples may introduce  unwanted noise and reduce the ratio of good samples.
Thus, a smaller value of $K$ may be more appropriate at the beginning.
Second, when the model gets to converge, the negative sampling ratio also needs to be smaller, and the optimal one is 1 when $\mu_{q(x)}=\mu_{q'(x)}$.
This is because easy samples are dominant when the numbers of good and bad samples are almost even, and focusing on optimizing the loss on easy samples may lead to overfitting.
Thus, a fixed negative sampling ratio may be suboptimal for  model training at different stages.

\section{ANS: Adaptive Negative Sampling}

Based on the findings in above section, we further propose an Adaptive Negative Sampling (ANS) method that can dynamically adjust the negative sampling ratio $K$ during model training to overcome the limitation of using a fixed one. 
We first introduce an extension of the standard negative sampling method where the negative sampling ratio $K$ is real-valued.
We denote the set of negative samples associated with each positive sample as $\mathcal{N}$, which satisfies the following formulas:
\begin{equation}\small
\begin{aligned}
    P(|\mathcal{N}|=[K])=1-\{K\},\\
     P(|\mathcal{N}|=[K]+1)=\{K\},\\
\end{aligned}
\end{equation}
where $[x]$ and $\{x\}$ represent the integral and fractional parts of $x$ respectively.

\begin{figure}[!t]
	\centering 
	\includegraphics[width=0.5\textwidth]{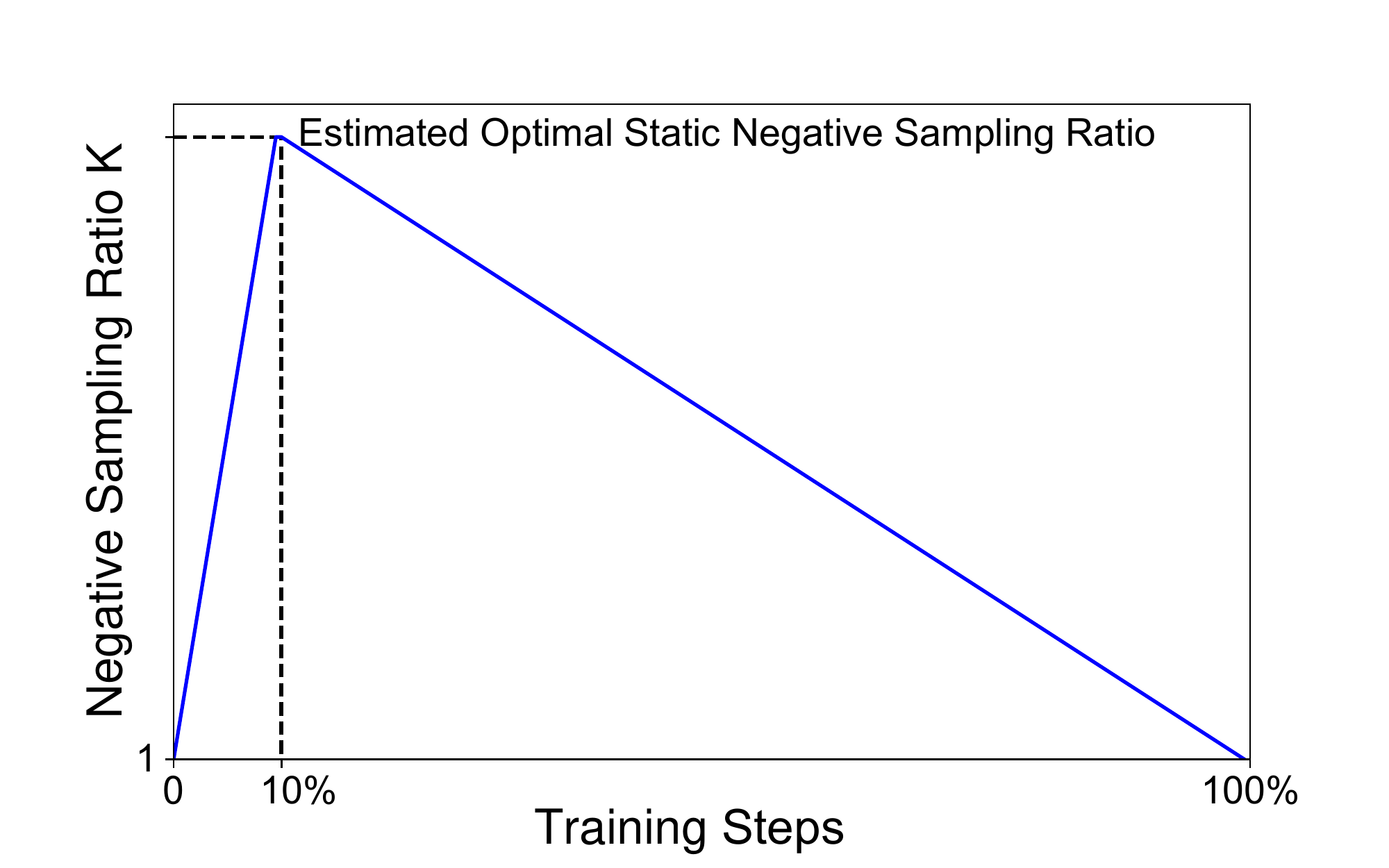}
	%\vspace{-0.05in}
\caption{The curve of negative sampling ratio in the  \textit{ANS} method.} \label{fig.ans}
%\vspace{-0.1in}
\end{figure}

We then introduce our proposed \textit{ANS} method.
Motivated by the analysis in the previous section and the popular learning rate warm-up strategy~\cite{goyal2017accurate}, we propose to adjust the negative sampling ratio $K$ as the curve shown in Fig.~\ref{fig.ans}.
The value of $K$ quickly grows from 1 to the optimal value estimated by our framework, and then slowly declines to 1 as the training continues.
This is because the optimal negative sampling ratio may be small at the beginning and the end of model training, and the performance of model usually improves quickly at the beginning.
The turning point on this curve is 10\% of the training steps, which is empirically selected based on experimental results (included in supplements).
Our \textit{ANS} method can adapt to different stages in the model training process, which can overcome the drawbacks of using a fixed negative sampling ratio.

\section{Experiments on the ANS Method}\label{sec:Experiments2}

\begin{figure}[!t]
	\centering 
	\subfigure[AUC.]{
	\includegraphics[width=0.4\textwidth]{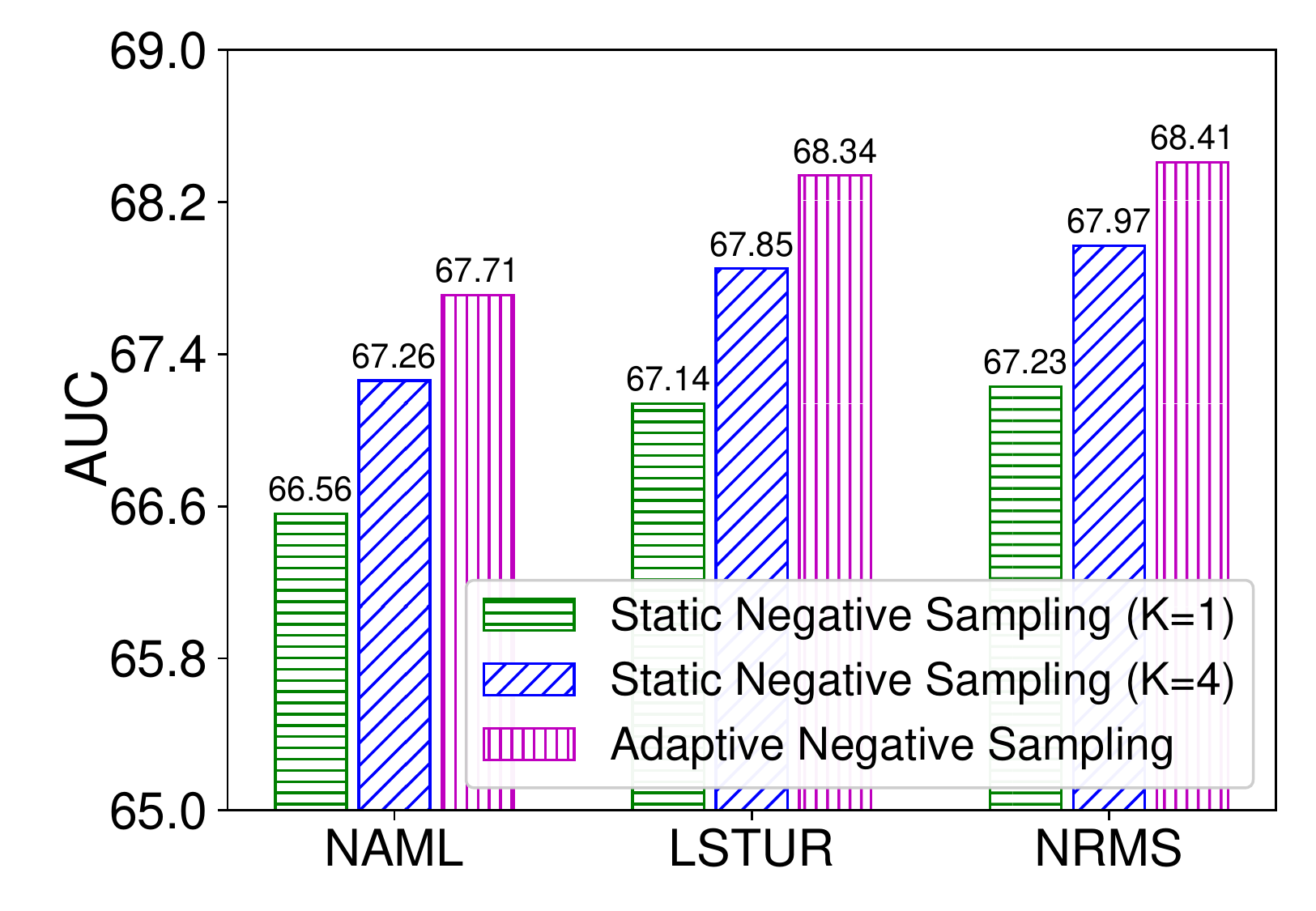} \label{fig.negnews1}}
 	\subfigure[nDCG@10.]{
	\includegraphics[width=0.4\textwidth]{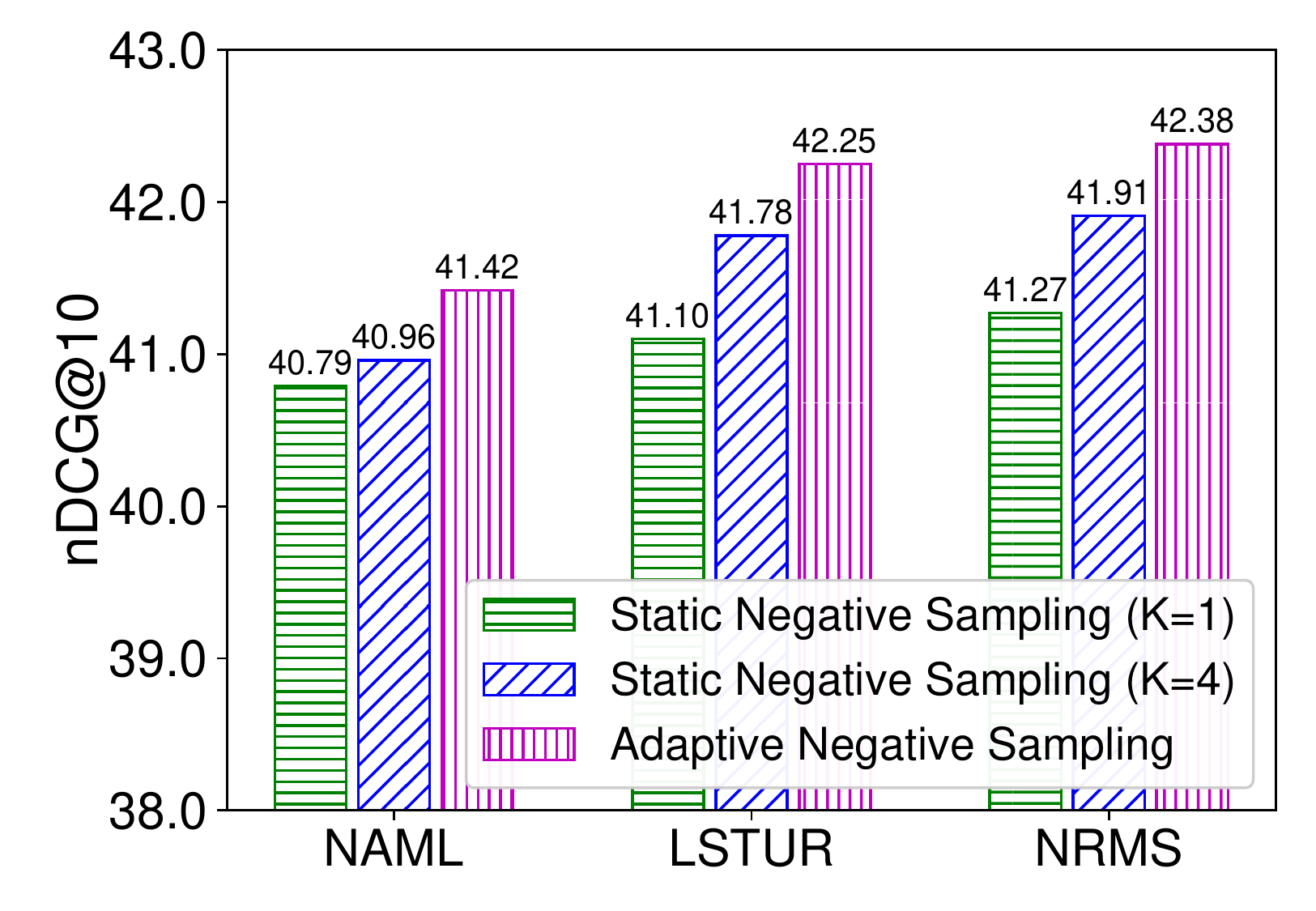} \label{fig.negnews2}}
	%\vspace{-0.06in}
\caption{News recommendation experiments with different negative sampling strategies. } \label{fig.negnews}
%\vspace{-0.12in}
\end{figure}

\begin{figure}[!t]
	\centering 
	\subfigure[AUC.]{
	\includegraphics[width=0.4\textwidth]{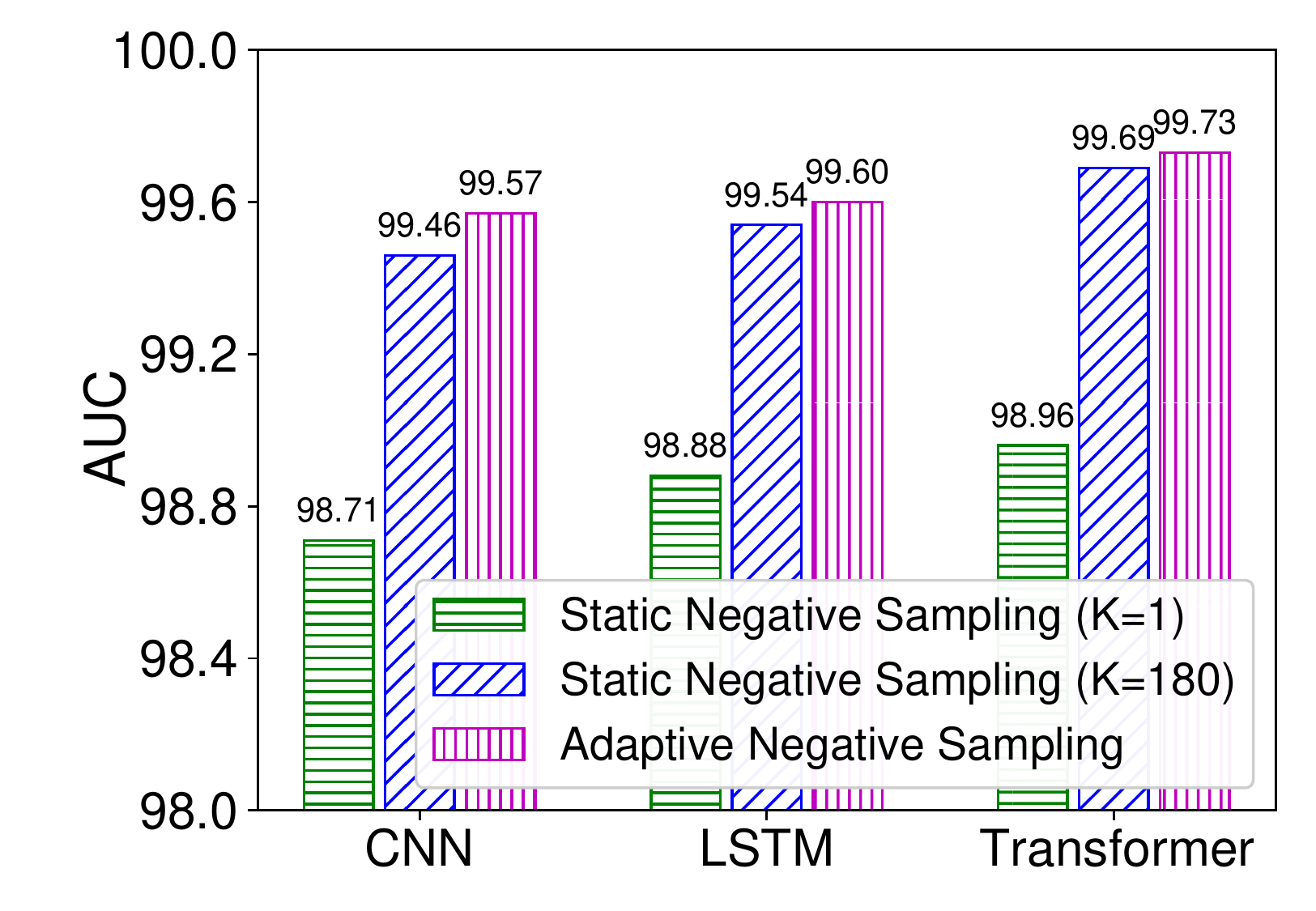} \label{fig.negmatch1}}
 	\subfigure[HR@5.]{
	\includegraphics[width=0.4\textwidth]{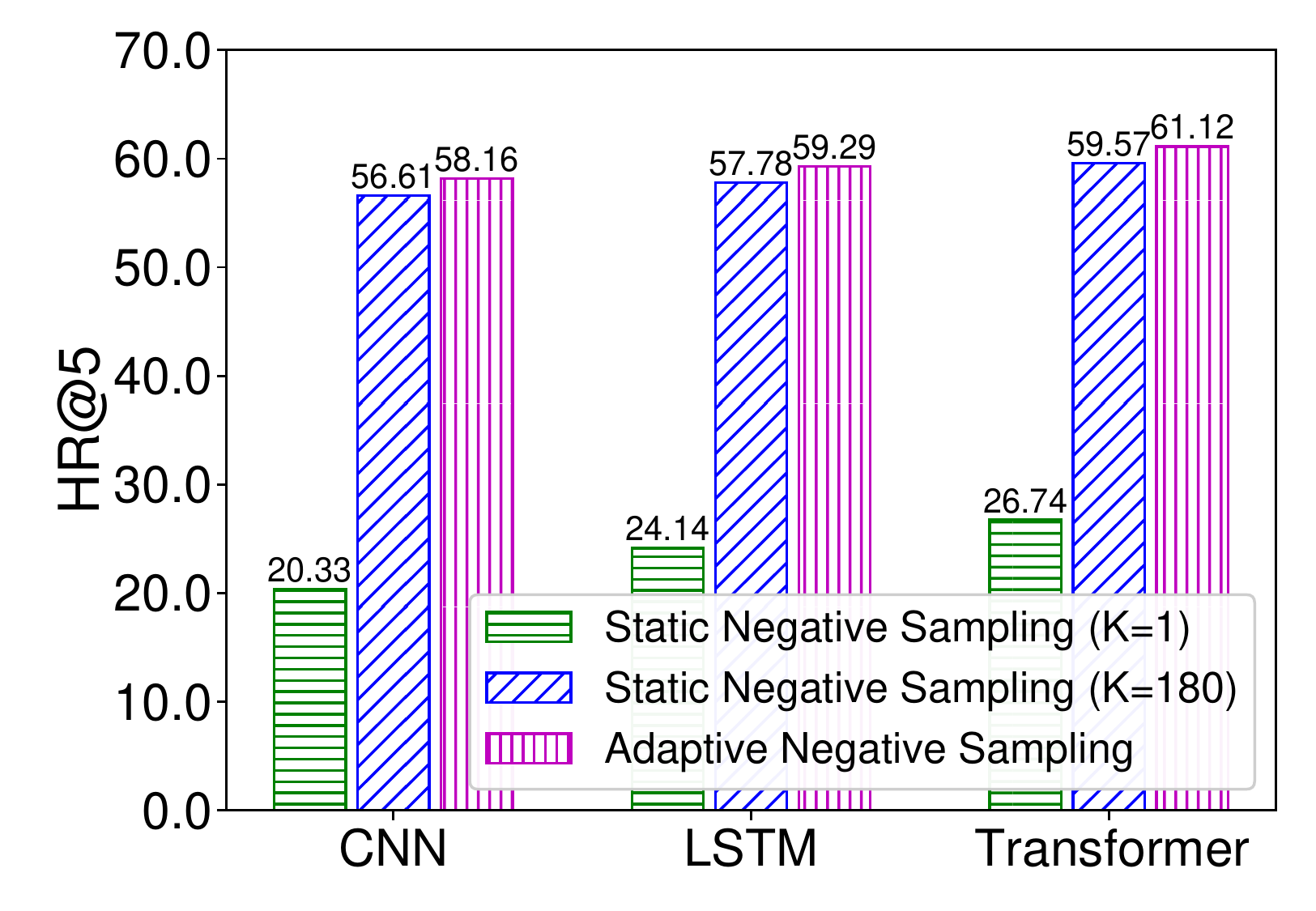} \label{fig.negmatch2}}
	%\vspace{-0.06in}
\caption{News title-body matching experiments with different negative sampling strategies. } \label{fig.negmatch}
%\vspace{-0.12in}
\end{figure}

In this section, we conduct experiments to verify the effectiveness of our proposed \textit{ANS} method.
On the news recommendation task, we apply our \textit{ANS} method to several state-of-the-art baseline methods, including NRMS~\cite{wu2019nrms}, LSTUR~\cite{an2019neural} and NAML~\cite{wu2019}.
On the news title-body matching task, we apply \textit{ANS} to the Siamese Transformer~\cite{reimers2019sentence} network and its variants based on LSTM or CNN.
We compare their performance with those trained with static negative sampling strategies that use a fixed negative sampling ratio.
The results are shown in Figs.~\ref{fig.negnews} and~\ref{fig.negmatch}.
We find that using the optimal negative sampling ratio estimated by our framework is better than popular default negative sampling ratio (e.g., 1)~\cite{rendle2009bpr}.
Moreover, using our proposed adaptive negative sampling method can achieve better performance than using static negative sampling ratio.
This is because \textit{ANS} can use different negative sampling ratios to better fit the characteristics of different training stages.

\section{Conclusion}\label{sec:Conclusion}

In this paper, we study how many negative samples are optimal for InfoNCE-based model learning in different tasks using a semi-quantitative theoretical framework.
We first propose a probabilistic model to analyze the influence of negative sampling ratio in InfoNCE on the informativeness of training samples.
Then, we propose a training effectiveness function to measure the overall influence of training samples on model learning based on their informativeness.
We further estimate the optimal value of $K$ that maximizes this measurement.
Based on our framework, we further propose an adaptive negative sampling method that can dynamically adjust the negative sampling ratio according to the characteristics of different model training stages.
We conduct extensive experiments on different real-world datasets for different tasks.
The results show that our framework can accurately estimate the optimal negative sampling ratio, and our adaptive negative sampling method can consistently outperform the commonly used fixed negative sampling ratio strategy.

Our work also has the following limitations.
First, in practice we need to first run experiments under $K=1$ to obtain the training and validation AUC curves to estimate the key parameters in our framework.
We will explore how to reduce the effort on estimating the optimal negative sampling ratio.
Second, we cannot obtain a closed-form solution of $\mu_{q(x)}$ and we need to numerically solve Eq.~(7).
We will work on this problem to obtain a closed-form solution.
Third, the turning point in our adaptive negative sampling method is empirically tuned rather than  automatically selected.
We will explore how to adaptively tune this hyperparameter in our future work. 
\section*{Broader Impact Statement}

In this paper, we introduce a semi-quantitative theoretical framework for estimating the optimal negative sampling ratio in  InfoNCE-based contrastive learning.
In addition, we propose an adaptive negative sampling method based on the findings of our theoretical framework.
Our work can be applied to various applications such as document retrieval, personalized recommendation and language model pre-training, to help estimate a proper number of negative samples and enhance model training.
There are many benefits to use our proposed approach, such as improving the model performance and reducing the effort on hyperparameter search.
However, there are also potential negative effects of using our approach:
(1) Due to the complexity of real-world applications (e.g., data distribution, task characteristic and model capacity), there may be some gaps between real experimental results and the predictions given by our theoretical framework, and the models may not yield the optimal performance.
(2) Our approach is not aware of the data biases, and the modeled trained with our recommended negative sampling ratio or our proposed \textit{ANS} method may inherit and even amplify these biases~\cite{chuang2020debiased}.   
We would encourage further work to understand and address the limitations and potential negative effects of our framework in specific applications.

\bibliography{emnlp2021}
\bibliographystyle{acl_natbib}

\clearpage

\section*{Supplementary Materials}

\subsection*{Relation between AUC and $\mu_{q(x)}$}

\begin{figure}[!h]
	\centering  
	\includegraphics[width=0.38\textwidth]{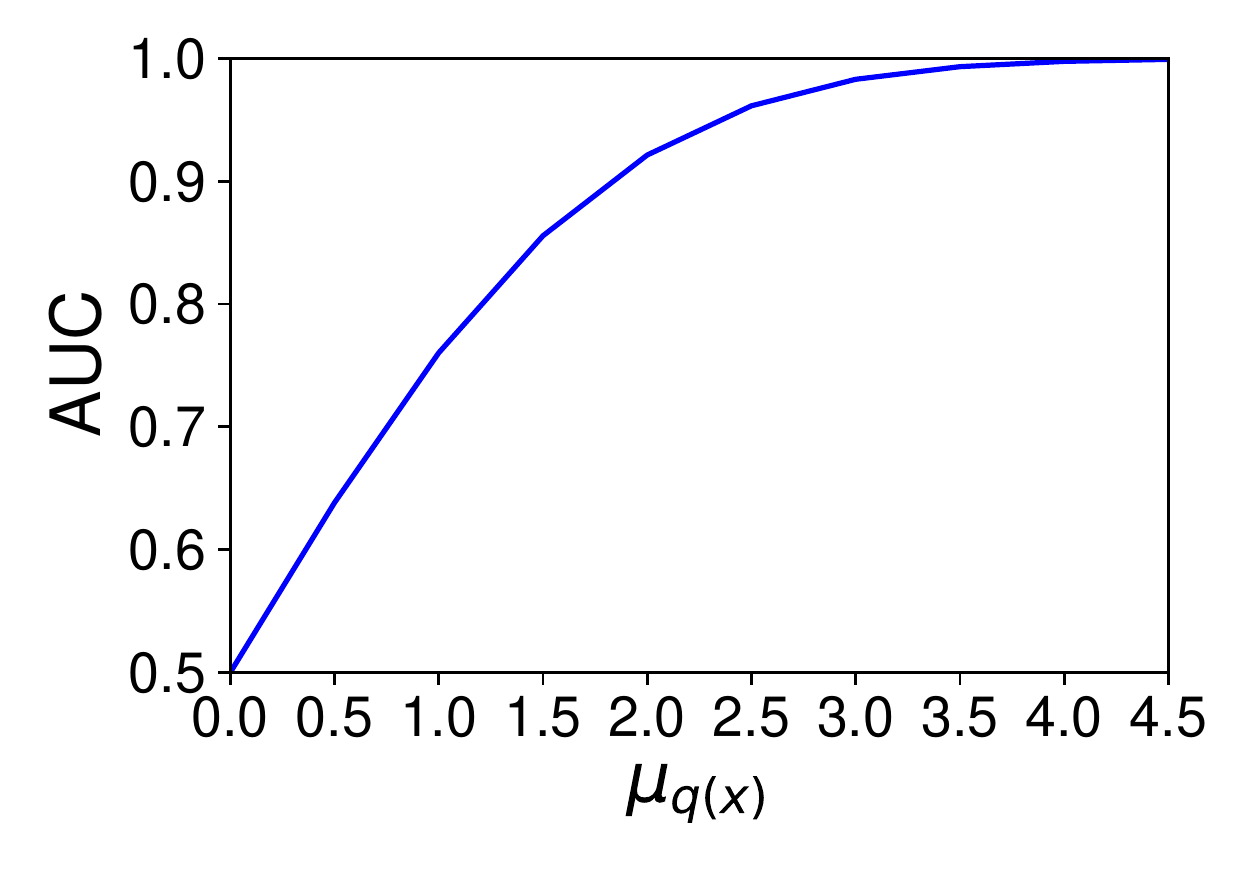} 
\caption{Relation between AUC and  $\mu_{q(x)}$.} \label{fig.aucmu}
\end{figure}

The relation between the AUC score and the value of $\mu_{q(x)}$ in our framework is shown in Fig.~\ref{fig.aucmu}.
We can see that the value of  $\mu_{q(x)}$  is 0 if AUC is 0.5, and increases when AUC gets close to 1.

\subsection*{Hyperparmeter Settings}

The detailed hyperparameter settings in our experiments are shown in Table~\ref{hyper}.

\begin{table}[!ht]
\centering
\caption{Detailed hyperparameter settings.}\label{hyper}

\resizebox{0.98\linewidth}{!}{
\begin{tabular}{|l|ccccccc|}
\hline
\multicolumn{1}{|c|}{Hyperparameters} & NRMS & LSTUR & NAML & \begin{tabular}[c]{@{}c@{}}Siamese\\ Transformer\end{tabular} & CNN  & LSTM & BERT4Rec \\ \hline
\# attention heads                    & 16   & -     & -    & 16                                                            & -    & -    & 2        \\
output head dim                       & 16   & -     & -    & 16                                                            & -    & -    & 32       \\
hidden dim                            & 256  & 256   & 256  & 256                                                           & 256  & 256  & 64       \\
dropout                               & 0.2  & 0.2   & 0.2  & 0.2                                                           & 0.2  & 0.2  & 0.2      \\
optimizer                             & Adam & Adam  & Adam & Adam                                                          & Adam & Adam & Adam     \\
learning rate                         & 1e-4 & 1e-4  & 1e-4 & 1e-4                                                          & 1e-4 & 1e-4 & 1e-4     \\
batch size                            & 32   & 32    & 32   & 32                                                            & 32   & 32   & 128      \\
total epoch                           & 2    & 3     & 2    & 5                                                             & 6    & 6    & 50       \\ \hline
\end{tabular}}

\end{table}

\subsection*{Implementation Details}
Several implementation details are introduced as follows.
We conducted experiments using a machine with Ubuntu 16.04 operating system and Python 3.6.
The machine has a memory of 256GB and a Tesla V100 GPU with 32GB memory.
We used the Keras  2.2.4 and tensorflow 1.12 to implement deep learning models.

\subsection*{Influence of Turning Point in ANS}

\begin{figure}[!t]
	\centering 
	\subfigure[News recommendation.]{
	\includegraphics[width=0.45\textwidth]{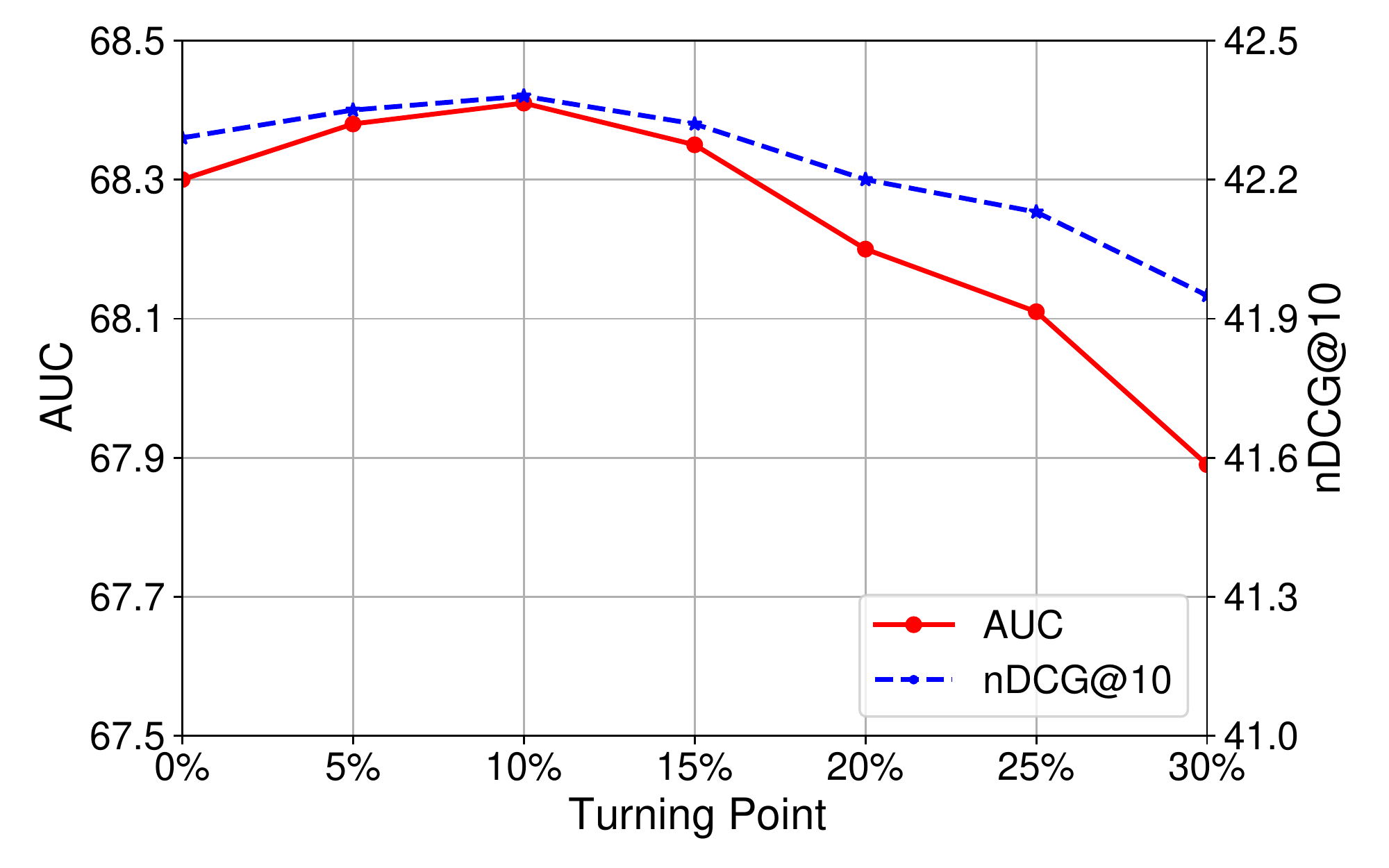} \label{fig.tip1}}
 	\subfigure[Title-body matching.]{
	\includegraphics[width=0.45\textwidth]{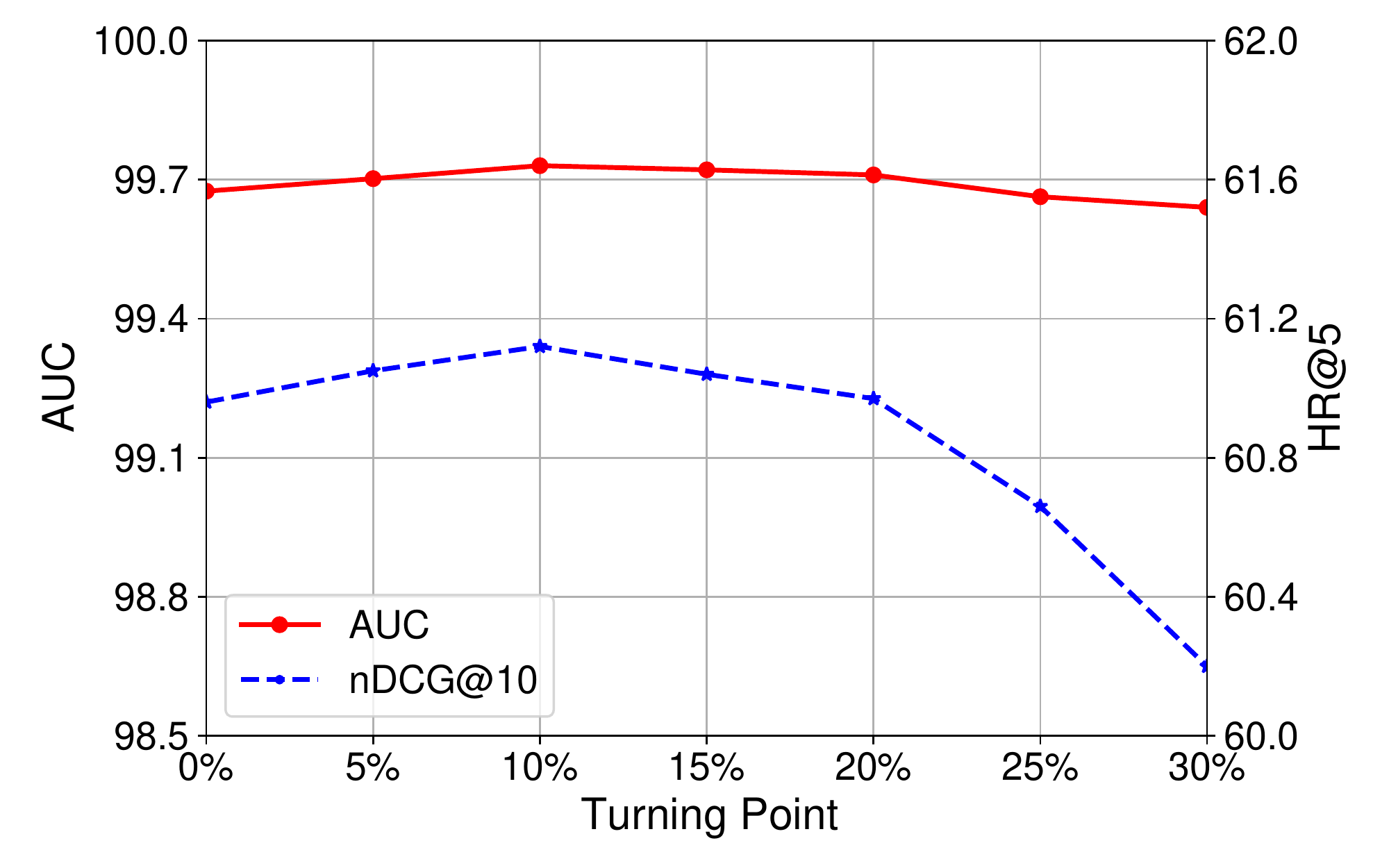} \label{fig.tip2}}
	\vspace{-0.06in}
\caption{Influence of turning points in our ANS method. } \label{fig.tip}
\vspace{-0.1in}
\end{figure}

We also conduct experiments to explore the influence of the turning point on the negative sampling ratio curve.
We compare the model performance by varying the turning point (using different percentages of iterations). 
The results are shown in Fig.~\ref{fig.tip}.
We observe that when the turning point appears too early, the performance is sub-optimal because the negative sampling ratio $K$ may need to be relatively small at the beginning.
In addition, if it appears too late, the performance is also sub-optimal because the model  may need a relatively larger $K$ after a certain number of iterations.
Thus, we empirically set the turning point to 10\% of the total training steps.

\end{document}